# Large Language Models in Fire Engineering: An Examination of Technical Questions Against Domain Knowledge

Haley Hostetter[1], M.Z. Naser[1,2], Xinyan Huang[3] John Gales[4]
[1]School of Civil and Environmental Engineering, and Earth Sciences, Clemson University, USA
[2]Artificial Intelligence Research Institute for Science and Engineering (AIRISE), Clemson University, USA
[3]Department of Building Environment and Energy Engineering, The Hong Kong Polytechnic University, Hong Kong
[4]Department of Civil Engineering, York University, Toronto, Canada
E-mail: hhostet@g.clemson.edu, mznaser@clemson.edu, xy.huang@polyu.edu.hk, jgales@yorku.ca

## Abstract
This communication presents preliminary findings from comparing two recent chatbots, OpenAI's *ChatGPT* and Google's *Bard*, in the context of fire engineering by evaluating their responses in handling fire safety-related queries. A diverse range of fire engineering questions and scenarios were created and examined, including structural fire design, fire prevention strategies, evacuation, building code compliance, and fire suppression systems (some of which resemble those commonly present in the Fire Protection exam (FPE)). The results reveal some key differences in the performance of the chatbots, with *ChatGPT* demonstrating a relatively superior performance. Then, this communication highlights the potential for chatbot technology to revolutionize fire engineering practices by providing instant access to critical information while outlining areas for further improvement and research. Evidently, and when it matures, this technology will likely be elemental to our engineers' practice and education.

## An introduction to Chatbots
A chatbot is a computer program capable of simulating conversation with humans [1]. Such a program (or platform) is often built via artificial intelligence (AI) and natural language processing (NLP) with the goal of understanding human questions and automating customized responses. More specifically, a chatbot is made from large language models (LLMs), a subset of AI trained via machine learning to learn the patterns and connections between words and phrases. This refers to the ability of the chatbot to uniquely respond to human-posed questions almost instantaneously, thereby simulating human conversation.

Perhaps the first idea for computer-based chatbots arose from the question, *"can a computer communicate in a way indistinguishable from human?"* posed by one of the computer science pioneers, Alan Turing, in 1950 [2]. In the decades following, several chatbots were developed, including *Eliza* (1964-6), *PARRY* (1972), *Racter* (1983), and *Dr. Sabaitso* (1991) [2]. Eliza, one of the oldest chatbots, was created by the Artificial Intelligence Laboratory at MIT and programmed to behave like a doctor—simulating the role of a psychotherapist who asks and responds to questions to divert attention back to the user [3]. *Eliza* quickly garnered the attention of users (who began to provide their personal stories, secrets, and sensitive information) and developers (who used *Eliza* as inspiration for later AI-powered chatbots). *PARRY*, on the other hand, was programmed to behave as a "paranoid schizophrenic patient" [4]. It attempted to provoke a user with questions that required more elaborate responses, thereby providing learning psychiatrists with a tool to help them communicate with schizophrenic patients. Later, *Racter* and *Dr. Sabaitso* increased the realism of chatbot technology. According to developer William Chamberlain, *Racter*'s ability to direct a computer to maintain certain randomly chosen variables, such as words and phrases, helps the program's answers pass for coherent thinking similar to humans [2]. Then, the chatbot *Dr. Sabaitso* (an acronym for "Sound Blaster Artificial Intelligent Text to Speech Operator") revolutionized the field as the first chatbot able to synthesize speech similar to a psychologist [5]. Although it could not communicate in a sufficiently complex way, verbal communication increased the humanity of the chatbot.

Apart from the aforementioned chatbots, the development and use of new AI chatbots increased in popularity with the expansion of the Internet and social media platforms in the 1990s. For example, *ALICE*, one of the first chatbots to go online, was introduced in 1995 [2]. An acronym meaning "Artificial Linguistic Internet Computer Entity," *ALICE*, though over twenty years old, remains usable today and is often credited as paving the way for newer chatbots [4].

More recently, chatbots have once again captured the interest of the general public and industries alike. This is primarily due to chatbots' ability to integrate into businesses, thus saving companies time and resources. Current chatbots, however, can have a variety of purposes and applications. For example, like the first chatbots, recent chatbots by *Disney* and *Marvel* are meant for entertainment purposes, allowing users to converse with their favorite movie characters [6,7]. Still others attempt to solve real-world problems or assist in the healthcare field (see the chatbot developed by the company *Endurance*, a companion for dementia and Alzheimer's patients [8] or the chatbot developed by the company *Casper*, a conversational chatbot for people with insomnia [9]).

With the broadening range of chatbot abilities and uses, many have questioned the possibility of using such AI for more scientific or practical applications or research. However, few papers have yet to address this, and available literature seems to focus on the education front. For example, Perez et al. [10] find using chatbots for education helpful—particularly for students with disabilities. They also note the possibility of bridging the educational "gap" between marginalized and mainstream groups. Similarly, Fryer et al. [11] designed an experiment to monitor students' learning behavior when their education was supplemented with a chatbot. Results found that chatbots can have a positive influence on student performance.

In more technical fields, such as medicine and engineering, extensive research on the applicability of chatbots is limited. For medicine, current research suggests promise for the application of chatbots, especially for automated and repetitive tasks [12]. Additionally, concerns about the legal and clinical aspects of using chatbots are common. In engineering, a 2022 study attempted to explore the implementation of chatbots into engineering education. Results show the possibility of using chatbot technology to assist in cooperation between students as well as the potential for their use as tools to support engineering design practice [13]. Prior to 2020, another study aimed to develop a chatbot capable of participating in the early stages of the engineering design process. Following interaction with several college students, the AI chatbot proved effective [14].

In the more niche field of fire engineering, experts seem to disagree with the aforementioned belief that chatbots are appropriate tools for education and design. One of the few available publications states their applicability in summarizing concepts and documents for non-experts, but that they cannot be used to replace or replicate fire engineers since such experts must ensure the safety of structures as well as conduct fire safety assessments, recommend design changes, and select fire protection systems [15].

With the present divided opinions on practical use and limited applicable literature in the engineering field, the authors aim to address chatbot responses and the possibility of their implementation into the fields of fire engineering and evacuation (including design and research). This is completed using the *Ask Me Anything* approach for two of the leading publicly available chatbots: OpenAI's ChatGPT [1,2] and Google's Bard. These two chatbots were released in November 2022 and February 2023 respectively, and our analysis examines the early release versions of these chatbots. This paper aims to explore the current capabilities of leading chatbots in a technical field and build upon the limited available research. With the fast-evolving fields of AI and machine learning, the authors comment on the current abilities of the chatbots in comparison to expert knowledge.

**Methodology**

This communication presents a comparison of two of the premier LLM chatbots (OpenAI's ChatGPT and Google's Bard) in the context of fire engineering by explicitly evaluating their accuracy in handling fire safety-related queries when displayed in the form of Ask Me Anything (AMA). A diverse range of fire engineering questions and scenarios were created, including structural fire design, fire prevention strategies, evacuation, building code compliance, and fire suppression systems (some of which resemble those commonly present in the Fire Protection exam (FPE)) [3]. The authors choose to qualitatively evaluate chatbot responses based on domain knowledge, published literature, and acceptable answers to possible questions that may resemble the FPE exam.

*Chatbot selection and definitions*

In late 2022, OpenAI released an early version of ChatGPT, a natural language processing tool driven by AI technology. By January 2023, ChatGPT had become the fastest-growing consumer software application in history, gaining over 100 million users in just two months [16]. To compete with ChatGPT, Google released its own chatbot, Bard, in March 2023. Both chatbots are generative AI platforms.

When these two chatbots were asked to define what a chatbot is by these authors, they returned the following definitions:

According to ChatGPT,

> A chatbot is an artificial intelligence (AI) software program designed to simulate human-like conversations with users, usually through text or voice interactions. Chatbots can understand and process natural language inputs, such as text messages or spoken words, and generate appropriate responses based on their

[1] In this paper, ChatGPT refers to ChatGPT-4, released in March 2023. For transparency, some of the answers might look different by the time this publication is released. This is due to the dynamic nature of these chatbots, as well as continued learning.

[2] One should note that, at the time of this paper's submission, ChatGPT did not have access to real-time online data (i.e., had a cutoff date to September 2021), unlike Bard. Microsoft Corporation launched an experimental model in March 2023 that used the Bing search engine to browse the Internet for more accurate data. This model was available only to paid subscribers of ChatGPT Plus [55].

[3] Recently, chatbots have been evaluated for domain-specific knowledge by comparing their answers to acceptable responses on standardized and professional licensing exams. Examples include [56,57]. Thus, we apply this concept to the fields of fire safety and fire protection engineering by evaluating ChatGPT and Bard's performances' when asked questions from the FPE exam.

> underlying algorithms and knowledge base. They are typically used to provide customer support, answer frequently asked questions, assist with various tasks, or offer personalized recommendations.

And according to Bard,

> A chatbot is a computer program that simulates human conversation. It can be used to answer questions, provide information, or perform tasks. Chatbots are often used in customer service applications, but they can also be used for a variety of other purposes.

In early 2023, ChatGPT and Bard were some of the only publicly available LLMs. With their rise in popularity worldwide, it is of interest to understand their capabilities and limitations in a more technical setting. Thus, we have selected both LLMs to evaluate their understanding of fire engineering.

*Question selection and answer evaluation*

Chatbots such as Bard and ChatGPT are intended to be open-ended devices that a user can access to ask any general or more detailed question they may wish. As such, the platforms lack a "standard" way in which to ask questions, formulate the structure of such questions, or determine the words to use in the queries. Thus, the questions posed to each chatbot in this communication were intended to mimic/replicate the questions a non-expert or student may ask about fire engineering or evacuation. We additionally ask questions that may arise in the Fire Protection exam for fire engineers and professionals. Thus, answers are evaluated based on domain knowledge, acceptable responses during the Fire Protection exam, and definitions and understanding provided in the open literature. Additionally, we aim to understand the present knowledge of such chatbots by posing *general* questions. We then evaluate the chatbots responses in terms of their understanding of minute differences that have been adopted in the practicing engineering field as well as academia.

*General chatbot architecture*

The architecture of a chatbot includes its underlying structure and design. It defines how a chatbot processes text and depends on a variety of factors such as domain, use-case, chatbot type, and many more. Thus, the key components of the chatbot architecture can be different for different bots. Currently, the most common types include *ruled-based, retrieval-based, generative, and hybrid.*

*Rule-based*, or *scripted*, chatbots are the earliest form of chatbots and are developed based on predefined rules. They follow such rules to generate responses using a series of conditional statements that check for keywords or phrases in a user's input. At the basic level, *rule-based* architecture includes three parts: a user interface (UI), Natural Language Processing (NLP) engine, and the rule engine [17]. The UI is the platform through which the user asks questions and views chatbot responses. The NLP engine works behind the UI to process the user's input and convert it to a readable format for the machine. Finally, the rule engine is responsible for interpreting the user's input, processing the input through the conditional rules of the chatbot, and returning the answer. *Rule-based* chatbots can be useful in certain scenarios but suffer from several limitations. These include a limited ability to understand natural language, a lack of contextual understanding, difficulty handling ambiguity and subtle nuances, and an inability to learn and adapt over time [17]. Additionally, *rule-based* chatbots are difficult to scale up and improve since new programming is required to update rules and patterns within the architecture.

Next, *retrieval-based* chatbots work by pulling chatbot responses from an established corpus of dialogs [18]. Such chatbots use machine learning models like supervised neural networks to interpret user inputs and determine fitting answers [18]. They rely on predefined responses (like *rule-based* chatbots) but can self-learn and improve their responses over time. Thus, these chatbots benefit from a greater ability to scale, adapt to changing user inputs and interactions, and require less maintenance throughout their lifetime. Additionally, given their predefined rules and responses, *retrieval-based* chatbots guarantee quality answers without grammatical errors. However, they suffer from many of the other same limitations of *rule-based* chatbots. They lack contextual and natural language understanding and have difficulty understanding nuanced inputs.

Third, *generative* chatbots are capable of formulating their own responses based on user input rather than relying on predefined rules or existing responses [18]. This requires the use of machine learning models like neural networks and large datasets to train the chatbots to make decisions about appropriate responses [19]. *Generative* models are powerful and flexible tools that benefit from an unconfined set of responses. However, their requirement of machine learning and large datasets make them challenging to implement. Additionally, it can be unclear how they make decisions and determine an appropriate answer for user input. This makes them more prone to grammatical errors and incorrect replies.

Finally, *hybrid* chatbots work by combining aspects of *rule-based, retrieval-based,* and *generative* chatbots [19]. They use a combination of pre-defined rules, pre-defined responses, and machine learning (neural networks) to deliver the best responses to user inputs. As a hybrid, they benefit from the same advantages of each of their components. They are scalable and have improved the quality of answers over generative models. However, they suffer from a limited ability to understand natural language and rely heavily on accurate data in their training sets [19].

*Rule-based, retrieval-based, generative, and hybrid* models each use NLP engines to process user input and convert it to machine-readable forms. However, as mentioned in the limitations of such models, NLP engines commonly lack the ability to understand natural human language. As a result, LLMs were developed with the intent to improve AI understanding of natural language and generate believable human text [20]. LLMs are a specific application of NLP and are created through a transformer model—a type of neural network (deep learning) architecture and form of generative AI. They work primarily through their transformer architecture and large training datasets and have been previously limited to use by large technology companies, who use them internally or on a limited basis [20]. However, LLMs are becoming increasingly available to the public and include models such as GPT created by OpenAI and BERT, LaMDA, and PaLM created by Google [20].

*ChatGPT* works under the *GPT,* or Generative Pre-trained Transformer, language model, while *Bard* works under the *BERT*, or Bidirectional Encoder Representations from Transformers, language model [20]. As their names imply, both *GPT* and *BERT* are transformer-based LLMs but work in different ways. GPT is an autoregressive LLM, meaning it uses past textual data and previous text inputs to determine the most appropriate next word or phrase to add to a sequence. *GPT* is built on a transformer *decoder* that allows individual outputs to be shared based on previously decoded outputs [20]. On the other hand, *BERT* is a collection of bidirectional language models from Google that have high levels of natural language and contextual understanding. It is built on a transformer *encoder,* so it generates and shares all of its outputs at once [20]. The difference in transformers generally means that *GPT* models are better at creating new human-like text, while *BERT* models are better at classification and summary tasks.

To be accurate, LLMs must be trained on large amounts of data. *ChatGPT* was trained using *Reinforcement Learning from Human Feedback* (RLHF) [21]. The model first went through a process of supervised fine-tuning, during which OpenAI trainers acted as both the human user and the AI bot. During this process, the trainers mimicked the way that humans communicate by creating a conversational sequence. Then, the dialogue was added to the model's dataset to improve it for conversational use. Later, the chatbot was improved using reinforced learning (i.e., rewarding the bot for generating correct responses and grading them from best to worst) [21]. Finally, OpenAI fine-tuned the model using its own Proximal Policy Optimization technique [21]. Bard's training process included using unsupervised learning techniques in which a large amount of unlabeled data was input into the model [22]. Then, the model processed the text to find statistical patterns and relationships, allowing Bard to generate appropriate responses for a wide variety of user inputs. Bard was also fine-tuned using supervised learning [22]. The size of *ChatGPT*'s training dataset was 175 billion parameters (for *GPT-3*) from five different sources while *Bard*'s was 1.56 trillion [23,24].

**Ask Me Anything (AMA): The series of questions and answers**

In this analysis, each chatbot was asked the same question and their answers were compared. We provide our commentary, as taken from leading fire-based organizations and literature, to most of the provided questions to assess the responses received from each chatbot. We first start with two historical questions; then, we dive into more technical questions. The presented questions stem from pure professional curiosity and tend to favor research and educational components. Toward the end of this analysis, we present five questions aimed at exploring the knowledge span, limitations, and possible imagination of ChatGPT and Bard. Overall, we present the outcome of 27 questions and answers.

*Question no. 1: What is fire engineering?*

According to the Society of Fire Protection Engineers (SFPE) [25], *fire protection engineering* is defined as:

> Fire protection engineering is the application of science and engineering principles. The principles when working with fire protection relate to protecting both people and the environment, built and natural, from the hazards of fire. This includes analyzing fire hazards, mitigating fire through design and construction of buildings, examination of building uses and industrial processes, and design, installation and maintenance of fire detection and suppression systems.

Additionally, *fire safety engineering* is defined generally according to BS 7974 [26] as:

> The application of scientific and engineering principles to the protection of people, property and the environment from fire.

Each of the chatbot's answers is listed herein:

OpenAI ChatGPT,

> Fire engineering is a multidisciplinary field that focuses on understanding fire behavior, prevention, protection, and suppression within the built environment. It involves the application of scientific and engineering principles to assess, design, and implement fire safety strategies in buildings, infrastructure, and industrial facilities.

Fire engineering aims to minimize the risk of fire, protect lives, property, and the environment, and ensure the continuity of operations in the event of a fire. It encompasses various aspects of fire safety, including:

- Fire prevention: Developing strategies to reduce the likelihood of fires, such as proper design, material selection, and adherence to building codes and regulations.
- Fire protection: Designing and implementing passive fire protection systems, such as fire-resistant construction materials, compartmentation, fire doors, and fire-resistant glazing, to prevent fire spread and maintain structural integrity during a fire.
- Fire detection and suppression: Designing and integrating active fire protection systems, such as fire alarms, sprinkler systems, and gaseous fire suppression systems, to detect and control fires at an early stage.
- Smoke control and management: Designing ventilation systems and smoke control measures to manage smoke spread.
- Egress and evacuation: Designing and planning safe and efficient means of escape, such as exit routes, emergency lighting, signage, and occupant load calculations to ensure that occupants can evacuate a building quickly and safely during a fire or other emergency.
- Human behavior in fire: Understanding how people react in fire situations, including factors like perception, decision-making, and movement patterns, which play a crucial role in developing effective evacuation strategies and fire safety systems.
- Fire risk assessment: Identifying potential fire hazards, evaluating their likelihood and consequences, and recommending appropriate mitigation measures to minimize risk.
- Performance-based design: Developing fire safety solutions based on performance objectives rather than prescriptive requirements, which allows for greater flexibility and innovation in design while still meeting fire safety goals.
- Fire investigation and analysis: Investigating the causes, origin, and development of fires, as well as analyzing their effects on structures, materials, and occupants, to improve fire safety knowledge and inform future designs.
- Codes, standards, and regulations: Developing, interpreting, and applying fire safety codes, standards, and regulations that govern the design, construction, and operation of buildings and facilities
- Education and training: Educating and training professionals, such as architects, engineers, and building officials, in fire safety principles and practices
- Firefighting tactics and strategy: Collaborating with fire service personnel to develop effective firefighting tactics, strategies, and incident command systems, taking into account the unique challenges posed by different types of structures and fire scenarios.
- 
- Interdisciplinary collaboration: Fire engineering often requires collaboration with other disciplines, such as architecture, structural engineering, mechanical engineering, electrical engineering, and urban planning, to develop comprehensive and effective fire safety solutions for various types of buildings and environments.
- Fire protection economics: Analyzing the costs and benefits of various fire protection strategies and systems to optimize resources and achieve the desired level of fire safety while balancing financial constraints.

## Google Bard

Fire engineering is the application of scientific and engineering principles to protect people, property, and their environments from the harmful and destructive effects of fire and smoke. It is a multidisciplinary field that encompasses the study of fire dynamics, fire suppression, fire protection systems, fire safety engineering, and fire investigation.

> Fire engineers use their knowledge of fire science to design and implement fire safety measures in buildings, industrial facilities, and other structures. They also develop and test new fire suppression technologies. Fire engineers are also involved in fire investigation, which is the process of determining the cause and origin of fires.
>
> Fire engineering is a critical part of the fire service. Fire engineers work with firefighters and other first responders to ensure that they have the tools and resources they need to protect the public from fire.
>
> Here are some of the tasks of a fire engineer:
>
> - Design and implement fire safety measures in buildings, industrial facilities, and other structures.
> - Develop and test new fire suppression technologies.
> - Conduct fire investigations.
> - Work with firefighters and other first responders to ensure that they have the tools and resources they need to protect the public from fire.

Question 1 was intended to test the chatbots' understanding of minute differences in the definitions of fire protection and safety engineering. Thus, the question was simply posed as "fire engineering". Generally, fire protection engineering refers to the implementation of prevention measures such as suppression and detection devices (sprinklers and alarms, respectively). On the other hand, fire safety engineering refers to the engineering design of a building or structure to prevent fire spread and/or loss of structural integrity. Neither Bard nor ChatGPT were able to communicate the difference between the two definitions. Instead, they combined the two. Given the general nature of the question posed to the chatbots, their answers seem acceptable. While ChatGPT provides a more detailed response, Bard's response is efficient. A fire engineer is a multidisciplinary engineer involved in a variety of fire-related scientific items (as detailed by ChatGPT). It would perhaps be helpful to understand, however, that fire engineers do not provide all the tasks detailed in the chatbots' responses in their specialized roles at one time. Instead, they may be involved with one or more engineers depending on their line of work.

*Question no. 2: Who was the world's first fire engineer?*

While the first fire engineer is hard to trace, the early account of firefighting is said to have begun in ancient Rome [27].

In terms of the examined chatbots,

OpenAI ChatGPT,

> It is challenging to pinpoint the world's first fire engineer, as the history of fire engineering spans across centuries and various cultures, with many individuals contributing to the development of fire safety practices and technologies over time.
>
> One of the early pioneers in the field of fire safety was Benjamin Franklin, who founded the Union Fire Company in Philadelphia in 1736, the first volunteer fire company in America. While Franklin was not a fire engineer in the modern sense, his contributions to fire safety and firefighting were crucial in laying the groundwork for the field.
>
> The modern concept of fire engineering began to take shape in the late 19th and early 20th centuries, as the industrial revolution and rapid urbanization led to new challenges in fire safety. Henry Grattan (H.G.) Tyrrell was an influential figure during this period. An American civil engineer, he published the book "Fire Engineering" in 1919, which was one of the first comprehensive texts on the subject.

> It is important to note that fire engineering has evolved through the collective efforts of many individuals, each contributing to the knowledge and practices we have today. While it is difficult to identify a single "first fire engineer," these early pioneers played a vital role in shaping the field.

Google Bard

> The world's first fire engineer was James Braidwood, a Scottish firefighter who was the first "Master of Engines", in the world's first municipal fire service in Edinburgh in 1824. He was the first director of the London Fire Engine Establishment (the brigade which was eventually to become the London Fire Brigade). He is credited with the development of the modern municipal fire service.
>
> Braidwood was born in Edinburgh in 1791. He joined the Edinburgh Fire Brigade in 1804 and quickly rose through the ranks. In 1824, he was appointed as the first "Master of Engines", a position responsible for the training and management of the fire brigade. Braidwood was a pioneer in the field of fire engineering. He developed new techniques for firefighting, such as the use of steam pumps and ladders. He also developed new fire prevention measures, such as the use of fire escapes and sprinkler systems. Braidwood died in 1861 while fighting a fire in London. He is considered to be one of the founding fathers of fire engineering.

Bard cites Wikipedia [28].

Overall, it is unrealistic to expect the chatbots to pinpoint a single person as the first fire engineer since we also collectively do not know the identity of who could be considered the first fire engineer. However, ChatGPT seems limited in its history of firefighting and engineering—listing only American history. Though Bard pinpoints an international fire fighter as the first fire engineer, the chatbot’s response is also limited by the timeframe. It seems that both ChatGPT and Bard are unable to recount history earlier than the Middle Ages.

Now we move toward more scientific questions.

*Question no. 3: Explain the process of fire growth in a compartment*

According to D.D. Drysdale in the Encyclopedia of Physical Science and Technology [29]:

> The term “compartment fire” is commonly used to describe a fire in a room or “compartment” within a building. If given sufficient fuel and ventilation, the fire will progress through three main stages:
>
> > 1. The growth, or preflashover, stage, when the fire is localized in the vicinity of its origin and the average temperature in the compartment is still relatively low.
> >
> > 2. The fully developed, or postflashover, fire, during which all the combustible items are involved and flames appear to fill the entire volume and emerge from the ventilation openings.
> >
> > 3. The decay period, when the fire begins to die down as the fuel is consumed. This is formally identified as the period after the average temperature has fallen to 80% of the maximum achieved during stage 2.
>
> The average temperature in the lower part of the compartment will remain low during the growth period, although a layer of hot smoke will accumulate under the ceiling and will increase in depth and temperature as the fire grows in size. The development of this layer is the key factor in the way in which a fire progresses to the fully developed stage. However, it is possible for a fire to self-extinguish if the item first ignited burns out before other items can become involved, or to be held in check if it becomes starved of air (oxygen). This can occur in a space where there is inadequate ventilation, as will occur if the doors and windows are closed and well fitting.

Then, the answers of each of the examined chatbots are listed herein.

OpenAI ChatGPT,

The process of fire growth in a compartment typically involves four stages: ignition, growth, fully developed, and decay. Here is a breakdown of each stage:

- Ignition: The ignition stage is the initial point at which the fuel source is ignited by an external heat source. This could be a spark, a flame, or any other ignition source that can provide enough heat to ignite the fuel.
- Growth: Once the fuel source has been ignited, the fire begins to grow rapidly. At this stage, the heat generated by the fire causes the surrounding fuel to heat up and release combustible gases, which in turn feed the fire and cause it to grow larger.
- Fully developed: In the fully developed stage, the fire has reached its maximum size and intensity. The heat generated by the fire is so intense that it begins to affect the surrounding air temperature and pressure, causing the fire to spread rapidly throughout the compartment.
- Decay: As the fuel source is consumed, the fire begins to slow down and eventually extinguishes. This stage is called the decay stage.

## Google Bard

Fire growth in a compartment is a complex process that is affected by a number of factors, including the type of fuel, the shape of the fuel, the ventilation conditions, and the presence of ignition sources.

The rate of fire growth in a compartment can be divided into three stages:

- The incipient stage is the initial stage of fire growth, when the fire is small and easily extinguished.
- The growth stage is the stage in which the fire grows rapidly and becomes more difficult to extinguish.
- The flashover stage is the stage in which the fire reaches its maximum intensity and becomes very difficult to extinguish.

The incipient stage is the most critical stage in fire growth, as it is the stage when the fire can be most easily extinguished. If the fire is not extinguished during this stage, it will continue to grow and become more difficult to extinguish.

The growth stage is the stage in which the fire grows rapidly and becomes more difficult to extinguish. During this stage, the fire will spread to new areas and become more difficult to contain.

The flashover stage is the stage in which the fire reaches its maximum intensity and becomes very difficult to extinguish. During this stage, the fire will produce a large amount of heat and smoke, making it difficult for firefighters to approach.

Both chatbots generally provide proper responses. ChatGPT's answer most closely matches the definition provided, only adding ignition. For Bard, the last stage was labeled the "flashover stage". In general, flashover is the point in time in which fuel is instantaneously ignited during a fire and occurs at approximately 600°C [30]. We generally consider flashover as the transition point between the growth and fully developed phases, not a stage itself. Despite the incorrect label of the phase, Bard's description is correct—a fully developed fire is one at its maximum intensity and is very hard to extinguish. However, Bard (as well as ChatGPT) fails to mention that a fire may become fully developed without reaching flashover, and that flashover will not always occur [31]. For flashover to occur, the fuel must have an adequate amount of heat energy and oxygen. For example, the presence of a closed door versus an open one in a room with a fire may be enough to prevent flashover due to the change in the amount of available oxygen (i.e. closed door provides less oxygen and thus reduces the possibility of flashover). Additionally, Bard did not include the decay phase of a compartment fire. A fire enters this stage as the available fuel is consumed or oxygen becomes limited. This may occur due if the compartment's ventilation is inadequate (i.e. the fire consumes more oxygen than can be replaced from the compartment's exterior) [31]. With decreasing oxygen, the heat release rate will also decrease [31]. Over time, the temperature also diminishes, although it may remain high for some time if the insulation is sufficient. Instead of including the decay phase, Bard instead commented on ways to reduce fire spread, which was not asked in the question.

*Question no. 4: Describe the mechanism(s) of fire-induced spalling of concrete*

The mechanisms behind fire-induced concrete spalling are complex and vary between four main theories: 1) pore pressure accumulation due to evaporating moisture, 2) the generation of thermal gradients, 3) the initiation of internal cracking due to difference in thermal expansion between concrete mixture components, 4) and temperature-induced degradation of strength loss [32,33].

Then, the answers of each of the examined chatbots are listed herein.

OpenAI ChatGPT,

> Fire-induced spalling of concrete occurs when the surface layers of concrete are exposed to high temperatures, causing the outer layer of the concrete to break away from the rest of the structure. The mechanism of spalling is complex and can be influenced by a variety of factors, including the type and quality of the concrete, the heating rate, the temperature distribution within the concrete, and the presence of moisture.
>
> There are several mechanisms that can contribute to fire-induced spalling of concrete:
>
> - Moisture-induced spalling: When concrete is exposed to high temperatures, the moisture within the concrete can evaporate rapidly, creating steam pressure within the concrete. If the steam pressure is not able to escape, it can build up and cause the concrete to crack and spall.
> - Thermal gradient-induced spalling: As the surface layer of the concrete is heated by the fire, it expands rapidly, creating thermal stresses within the concrete. If the thermal stresses exceed the tensile strength of the concrete, it can crack and spall.
> - Chemical decomposition-induced spalling: High temperatures can cause chemical reactions within the concrete that can weaken the bonds between the cement paste and the aggregates. This can lead to the formation of voids within the concrete, which can cause it to crack and spall.
> - Mechanical damage-induced spalling: The stresses caused by the expansion of steel reinforcement bars within the concrete can cause it to crack and spall. This is particularly true if the steel reinforcement has been corroded due to exposure to moisture.

Google Bard

> Fire-induced spalling of concrete is a complex phenomenon that is not fully understood. However, it is known that the following mechanisms can contribute to spalling:
>
> - Temperature-induced expansion and contraction. When concrete is heated, it expands. When it cools, it contracts. This expansion and contraction can cause the concrete to crack.
> - Pore pressure buildup. When water evaporates from concrete, it leaves behind empty pores. These pores can fill with water vapor, which can increase the pressure inside the concrete. This pressure can cause the concrete to crack.
> - Chemical reactions. When concrete is exposed to high temperatures, it can undergo chemical reactions that can weaken the concrete. These reactions can also cause the concrete to crack.
> - Micro-cracking. Even before fire-induced spalling occurs, concrete can undergo micro-cracking. These cracks are too small to be seen with the naked eye, but they can weaken the concrete. When fire-induced spalling occurs, these micro-cracks can grow larger and cause the concrete to spall.

Both chatbots have precise responses to this question. Spalling is complex phenomenon that takes place in concrete exposed to high temperatures. In addition to listing the mechanisms of spalling, ChatGPT additionally defines spalling while Bard does not.

*Question no. 5: What is the concept of fire equivalence?*
The concept of fire equivalence aims to establish a comparison between two different fires. Most notably, a real fire is hoped to be compared to a standard fire [34,35].

In terms of the examined chatbots,

OpenAI ChatGPT,

> The concept of fire equivalence, also known as fire load or heat release rate equivalence, is a method used to compare the potential fire hazard of different materials, objects, or spaces within a building. It provides a common basis for understanding the potential heat release and energy contribution of different combustible materials during a fire.
>
> Fire equivalence is typically expressed in terms of the potential heat release per unit area or volume (e.g., $MJ/m^2$ or $MJ/m^3$). It takes into account the mass of the combustible materials, their heat of combustion, and the available surface area for combustion. By comparing the fire equivalence values of different materials or spaces, it is possible to assess their relative fire hazards and inform decisions about fire protection strategies and building design.

Google Bard

> The concept of fire equivalence is a way of comparing the severity of different fires. It is based on the idea that two fires that have the same maximum temperature will have the same impact on a structure.
>
> Fire equivalence is typically measured in terms of the time it takes for a fire to reach a certain maximum temperature. For example, a fire that reaches a maximum temperature of 1,000 degrees Celsius in 10 minutes is said to have a fire equivalence of 10 minutes.
>
> Fire equivalence is a useful tool for fire engineers, as it allows them to compare the severity of different fires and to design structures that are able to withstand fires of a certain equivalence.
>
> However, it is important to note that fire equivalence is not a perfect measure. The actual impact of a fire on a structure will depend on a number of factors, including the type of structure, the fuel load, and the ventilation conditions.

Bard cites Wikipedia [28]

As one can see, both chatbots managed to relate the equivalence concept to a comparison between two fires. However, both chatbots also do not seem to fully grasp this concept. For example, ChatGPT presents ".. understanding the *potential heat release* and *energy contribution* of different combustible materials during a fire..". On the other hand, Bard relates the equivalence to attaining a specific temperature of ".. reach a *certain maximum temperature* ..". It is well known that tying equivalence to a maximum temperature is one method of establishing equivalence. However, this is not the only method.

*Question no. 6: In terms of beams and columns, what is the best geometric shape that limits heat transfer?*
The work of Kodur and Mackinnon [36] has shown that circular shapes are known to have lower heat transfer than edged shapes.

OpenAI ChatGPT,

> In general, hollow or tubular shapes can provide better thermal insulation than solid shapes because they create an insulating air gap. This reduces the overall heat transfer through the structural element. Circular or rectangular hollow sections are commonly used as columns, while hollow I-beams or hollow rectangular beams can be used for beams.

Google Bard

> The best geometric shape for beams and columns in terms of heat transfer is a square or rectangular cross-section. This is because the corners of a square or rectangle are rounded, which reduces the amount of surface area that is exposed to the air. This in turn reduces the amount of heat that can be lost from the beam or column.

A follow-up question was aimed at Bard, *I think the best shape would be a circular cross section.*

> A circular cross-section is not the best shape for beams and columns in terms of heat transfer. A circular cross-section has a larger surface area than a square or rectangular cross-section, which means that it will lose heat more quickly. In addition, a circular cross-section is not as efficient as a square or rectangular cross-section in terms of building structures.
>
> However, a circular cross-section is the best shape for beams and columns in terms of strength. A circular cross-section is the strongest shape because it has the least amount of material for the same amount of volume.

ChatGPT provides the correct answer to this questions, while Bard's answer is incorrect. Bard states that the best geometric shape for heat transfer is a square or rectangular cross-section because "the corners of a square or rectangle are *rounded."* Clearly this is not correct. Even with a follow up question, the chatbot is still incorrect and states that square/rectangular sections lose heat more quickly due to their larger surface area.

*Question no. 7: What is the expected fire resistance in terms of hours of a square reinforced concrete column with a width of 400 mm and a length of 4 m?*

According to Eurocode 2 [37] (Tables 5.2a and b), the expected fire resistance of a square reinforced concrete column with a width of 400 mm and a length of 4 m that is exposed on more than one side can vary between 90-120 min.

Then, the answers of each of the examined chatbots are listed herein.

OpenAI ChatGPT,

> The expected fire resistance of a square reinforced concrete column with a width of 400 mm and a length of 4 m will depend on several factors, including the thickness and quality of the concrete, the size and spacing of the reinforcement bars, and the type and duration of the fire exposure.
>
> Based on typical fire resistance ratings for reinforced concrete columns, a square reinforced concrete column with a width of 400 mm and a length of 4 m would be expected to have a fire resistance rating of approximately 2 to 4 hours. However, it is important to note that the actual fire resistance may vary depending on the specific design and construction details of the column, as well as the type and duration of the fire exposure.

Google Bard

> The expected fire resistance of a square reinforced concrete column with a width of 400 mm and a length of 4 m is 2 hours.

Here, Bard has the more correct answer. 2 hours is the high-end fire resistance for the described column. We can consider ChatGPT's response incorrect since 2-4 hours greatly exceed the expected fire resistance according to Eurocode. If one were to rely on this information, the column would actually fail much before specified by ChatGPT, and thus could cause structural damage and potential life safety issues.

*Question no. 8: What is the percentage of cost associated with fire design in terms of a typical building's design?*
Common practice dictates that the answer is between 3-5%.

Then, the answers of each of the examined chatbots are listed herein.

OpenAI ChatGPT,

> The percentage of cost associated with fire design in a typical building's design can vary widely depending on various factors such as the type of building, its size, and the complexity of the fire safety system. However, generally, the cost of fire design typically ranges from 2% to 5% of the total construction cost.
>
> While fire design may represent a relatively small percentage of the total construction cost, it is a critical component of building design that can have a significant impact on the safety of occupants and the protection of property. Investing in high-quality fire design can help to reduce the risk of fire-related damage, minimize insurance costs, and ensure the long-term safety and viability of the building.

Google Bard

> The cost of fire design is typically 2-5% of the total cost of a building's construction. This means that for a $100 million building, the cost of fire design would be $2-5 million.
>
> The cost of fire design is an important part of the total cost of a building's construction. By investing in fire design, building owners can help to ensure that their buildings are safe and compliant with fire codes.

This question refers to the construction and fabrication cost of fire protection and suppression systems as a portion of total construction costs. Both chatbots' answers match our domain knowledge. It is also worth noting that the "cost of fire design" may also refer to the cost associated with the analysis and design of items such as fire loading, structural design (including but not limited to beam/column design, passive protection methods, etc. for fire) by engineers as office or desk-based work, although this was not the intention of the question. Common practice dictates this to be between 5-15% of total engineering design cost. Neither chatbot mentioned this type of cost.

*Question no. 9: What is fire evacuation?*
According to Cambridge English Dictionary [38], evacuation is:

> "the act of moving people from a dangerous place to somewhere safe"

Similarly, emergency evacuation is the urgent, immediate egress or escape of people away from an area that contains an immediate threat to lives or property. Evacuation can be small- or large-scale and due to a variety of reasons (including but not limited to fire, etc.). In general, evacuation can be divided into two main phases: pre-evacuation and movement. In the pre-evacuation phase, evacuees gather personal items and make decisions about how they will evacuate from danger.

This phase includes detection, decision, alarm, and reaction. In the movement phase, evacuees physically move toward safety using the decisions they made in the previous phase [39].

OpenAI ChatGPT,

> Fire evacuation is the process of safely and efficiently moving people away from a building, structure, or area where a fire is present or has the potential to occur. The primary goal of fire evacuation is to protect the lives of the occupants by minimizing injuries, fatalities, and exposure to toxic smoke and gases.
>
> Fire evacuation procedures typically involve several key components:
>
> 1. Early detection and warning: Fire alarms and smoke detectors play a crucial role in alerting occupants of a fire and initiating the evacuation process.
> 2. Evacuation routes and exits: Pre-determined evacuation routes and clearly marked exits should be established in a building to guide occupants to safety. These routes should be free from obstructions and provide the shortest, safest path to the outside.
> 3. Fire drills and training: Regular fire drills and safety training help familiarize building occupants with evacuation procedures, ensuring that they know how to respond quickly and calmly in the event of a fire.
> 4. Assembly points: Designated assembly areas outside the building should be established, allowing evacuees to gather and be accounted for, and ensuring that emergency responders have clear access to the building.
> 5. Emergency communication: Effective communication during an evacuation is essential for relaying important information, such as the location of the fire or any changes in the evacuation plan.
>
> Fire evacuation plans should be tailored to the specific needs of the building and its occupants, taking into consideration factors such as building layout, occupancy levels, and the presence of individuals with disabilities or other special needs.

Google Bard

> A fire evacuation is the process of removing people from a building or other structure that is on fire. The goal of a fire evacuation is to ensure the safety of all occupants by removing them from the fire and smoke as quickly as possible. Fire evacuations are typically carried out by trained professionals, such as firefighters or building security personnel. However, it is important for all occupants of a building to be familiar with the fire evacuation plan and to know how to exit the building safely in the event of a fire.
>
> The following are some important steps to take in the event of a fire evacuation:
>
> - Stay calm. Panicking can lead to injuries and can make it difficult to evacuate the building safely.
> - Follow the instructions of the fire department or other emergency personnel. They are trained to ensure the safety of all occupants and will direct you to the safest exit route.
> - If you are unable to evacuate on your own, call for help. The fire department will be able to provide assistance and help you to evacuate the building safely.
> - Once you have evacuated the building, stay away from the area until it is safe to return. The fire department will need time to extinguish the fire and to ensure that the area is safe for occupants to return.
>
> Fire evacuations are a serious matter, but by following the steps above, you can help to ensure your safety and the safety of others in the event of a fire.

ChatGPT and Bard both provide valid responses to this question. An evacuation can be prompted for a variety of scenarios and involve many different methods of reaching safety. Most importantly, it is the movement of people away from an area of danger. Additionally, Bard is correct in noting that evacuation can be planned or unplanned. However, a planned evacuation does not constitute a well-organized one or one without urgency/chaos. It is important to note that evacuation does not require a mechanical method of transportation, as Bard describes (buses, boats, helicopters, etc.). Evacuation can also describe the movement of people by walking/running, such as when building occupants evacuate during a fire or fire drill. We would like to point out that Proulx [40] and Fahy [41] offered explanations for debunking the myth of panic through stress models. A concept that has been clearly adopted by Bard.

*Question no. 10: Who is the most vulnerable type of person for injury/death in a building fire?*
According to researchers from the National Institute of Standards and Technology [42], older persons (elderly), young children, and males are some of the most vulnerable people to fire injury and death.

OpenAI ChatGPT,

> During a building fire, certain groups of people are more vulnerable to injury or death due to factors such as limited mobility, cognitive impairments, or other specific needs. The most vulnerable groups include:
>
> - Elderly individuals: Older people may have reduced mobility, slower reaction times, and age-related health issues, which can hinder their ability to evacuate quickly and safely.
> - Children: Young children, particularly those below the age of five, are at greater risk due to their limited understanding of danger, inability to follow complex instructions, and dependence on adult caregivers for guidance and protection.
> - People with disabilities: Individuals with physical, sensory, or cognitive disabilities may face challenges during evacuation, such as navigating through a building, understanding alarms, or communicating with others.
> - People with chronic illnesses or medical conditions: Those with chronic health issues may have reduced stamina, limited mobility, or other factors that hinder their ability to evacuate efficiently.
> - Pregnant women: Pregnancy may limit mobility, stamina, or the ability to navigate obstacles, increasing the risk of injury or death during a building fire.
> - Non-native speakers or those with limited language proficiency: People who do not understand the local language may have difficulty comprehending evacuation instructions, announcements, or other critical information, which can impede their ability to evacuate safely.

Google Bard

> The most vulnerable type of person for injury/death in a building fire is children. Children are more likely to be injured or killed in a fire because they are:
>
> - More likely to be home alone when a fire starts
> - Less likely to know how to escape a fire
> - More likely to be curious and explore, which could lead them into dangerous areas
> - More likely to be unable to self-rescue
>
> Children are also more likely to be injured or killed in a fire because they have a smaller body surface area, which means they absorb heat more quickly. They also have a higher respiratory rate, which means they inhale smoke and fumes more easily.

ChatGPT seems to have answered this question thoroughly, even pointing out that non-native speakers and pregnant women are at additional risk for injury/death in building fires. Bard, on the other hand, only identified one type of person at risk during fires: children. While children are at increased risk during fires, they are not necessarily the most vulnerable. This response fails to include the elderly and people with disabilities, who are most commonly cited as having the most limitations during evacuations.

*Question no. 11: What are the two primary phases of a building evacuation?*

According to Rahouti et. al. [43]:

> The engineering evacuation timeline divides the evacuation process broadly into two evacuation performance phases following on from the fire detection and notification phases: pre-evacuation and evacuation movement. The pre-evacuation stage extends from the notification of the population (i.e. the point at which they might be aware of the incident) to the moment when an evacuee starts moving towards a place of safety; the movement phase starts directly after the pre-evacuation phase, and ends once the evacuee has reached a place of safety.

OpenAI ChatGPT,

> A building evacuation typically consists of two primary phases:
>
> - Alert and notification phase: This phase involves detecting the emergency, such as a fire or a security threat, and activating the alarm system. In this stage, occupants are alerted to the emergency, and the appropriate authorities, like the fire department or police, are notified. The alarm system may include visual and audible signals, like flashing lights and sirens, to ensure that all building occupants, including those with hearing or visual impairments, are informed of the need to evacuate. Clear communication of the emergency situation and evacuation instructions is crucial during this phase.
> - Evacuation and movement phase: This phase involves the actual process of moving occupants from the building to a safe location outside. It starts with occupants leaving their immediate area and progresses to them following designated evacuation routes towards exits. The evacuation process may be structured differently depending on the building's layout and the nature of the emergency. For example, a phased evacuation could be implemented for large, multi-story buildings, where certain floors or sections are evacuated first to avoid congestion and bottlenecks. In some cases, a shelter-in-place strategy might be used if evacuation is deemed more dangerous than staying inside the building. This phase also includes accounting for all occupants at the assembly point and ensuring that everyone has safely evacuated.

Google Bard

> The two primary phases of a building evacuation are:
>
> - Alarm and notification: This phase involves alerting people to the danger and instructing them to evacuate. This can be done through a variety of methods, such as using a public address system, ringing bells, or using flashing lights.
> - Evacuation: This phase involves people leaving the building and going to a safe location. This can be done through a variety of methods, such as using stairs, elevators, or ladders.

Both chatbots have responded correctly to this question. In general, evacuation consists of a pre-evacuation phase and a movement phase. Additionally, both platforms mention a broader plan for emergency response that includes pre-planning, clear communication, and regular practice; while this does not directly answer the question, it is an appreciated addition. Both chatbots label the first phase the “alarm and notification phase”. This is mostly correct, given that building occupants are generally notified of an emergency via fire alarm (or another type). However, it is important to note the differences in how each chatbot described the phase. For ChatGPT, the alarm and notification stage is described in actions the evacuees must go through prior to the evacuation or movement phase. For Bard, the phase was described not for the evacuees but instead for fire safety officials or others who can alert evacuees to danger. In an evacuation, the pre-movement phase most closely aligns with the evacuees' actions, not the fire safety officials.

*Question no. 12: Why are people with disabilities more vulnerable than those without disabilities in building fires?*

In general, people with disabilities are more vulnerable than those without disabilities in building fires because they experience limited mobility or difficulty in understanding the severity of the fire scenario (i.e., people with sensory disabilities may not hear the fire alarm, and those with cognitive disabilities may not be able to make informed decisions about the fire). Thus, they may move slower, have difficulty traversing some egress components such as stairs, or react slower than the non-disabled population. Ultimately, their rate of evacuation may be too slow compared to the rate of fire spread and result in injury or death. Additionally, egress routes and evacuation considerations are rarely geared toward people with disabilities [44]. This may increase their vulnerability in emergency situations.

OpenAI ChatGPT,

> People with disabilities are often more vulnerable than those without disabilities during building fires for several reasons:
>
> - Limited mobility: Individuals with physical disabilities, such as those who use wheelchairs, walkers, or crutches, may face challenges navigating through narrow corridors, stairwells, or uneven surfaces during an evacuation. Their movement speed may also be slower, which can put them at a greater risk in a rapidly developing fire situation.
> - Sensory impairments: People with visual or hearing impairments may have difficulty receiving or interpreting fire alarms and emergency instructions. For instance, a person with a hearing impairment might not hear an audible alarm, while someone with a visual impairment might struggle to navigate through a smoke-filled environment or read emergency exit signs.
> - Cognitive impairments: Individuals with cognitive disabilities, such as intellectual disabilities or dementia, may have difficulty understanding the urgency of a fire situation, following evacuation instructions, or making quick decisions during an emergency.
> - Communication barriers: People with speech impairments or language disorders might have trouble communicating their needs or location to emergency responders, which could hinder their rescue or evacuation.
> - Dependence on assistive devices or personal assistance: Some individuals with disabilities rely on assistive devices, such as wheelchairs or hearing aids, or require personal assistance from caregivers during daily activities. In an emergency, these individuals may need extra time or assistance to evacuate, which can increase their vulnerability.

Google Bard

> People with disabilities are more vulnerable than those without disabilities in building fires for a number of reasons:
>
> - They may have difficulty escaping a fire. People with disabilities may have difficulty using stairs or elevators, and they may not be able to move quickly enough to escape a fire.
> - They may have difficulty hearing or seeing a fire. People with disabilities may not be able to hear the smoke alarm or see the smoke or flames.
> - They may have difficulty communicating with others. People with disabilities may not be able to communicate with others in an emergency situation.

- They may have difficulty accessing emergency services. People with disabilities may have difficulty accessing emergency services, such as calling 911 or getting to a safe location.

People with disabilities other than physical disabilities are rarely studied in terms of fire evacuation [44]. Thus, ChatGPT's inclusion of people with sensory and cognitive disabilities is an important inclusion in its response of this question. ChatGPT further includes those with physical disabilities and communication barriers. Each item discussed can play a significant role in evacuation and fire vulnerability and are not present for the non-disabled population. Bard presents a more general answer but explains how each factor can affect the vulnerability of people with disabilities when compared to those without. Each response is deemed good. We would like to note that the specific US-based fire department (i.e., 911), as opposed to 999, is often used in other countries.

*Question no. 13: How can researchers study evacuation?*

In the past, researchers most commonly used real-time evacuation experiments and historical events (such as the MGM Grand fire [45] or 9/11 evacuation [46]) to study egress and human behavior. More recently, evacuation simulation and modeling have been introduced as a cost- and time-effective alternative [47–49]. This method is more ethical, does not pose safety risks to participants, and does not require a real building.

OpenAI ChatGPT,

Researchers can study evacuation without putting people at risk or relying on actual natural disasters by using various methods, including simulations, virtual reality, mathematical modeling, and controlled experiments. These approaches help researchers understand human behavior, assess evacuation procedures, and improve safety measures. Some methods used to study evacuations are:

- Computer simulations: Researchers use software to simulate evacuations, considering factors like building layouts, exit locations, and human behavior. By adjusting these parameters, they can analyze how different scenarios impact evacuation efficiency and identify potential bottlenecks or hazards.
- Virtual reality: Virtual reality (VR) provides a safe and immersive environment to study evacuation behavior. Participants equipped with VR headsets experience realistic emergency situations, allowing researchers to observe their decision-making and movement patterns without exposing them to real danger.
- Agent-based modeling: This approach involves creating computational models of individual agents (people) within a simulated environment. Researchers can assign specific characteristics to these agents, such as speed, decision-making abilities, and group dynamics, to understand how they collectively impact evacuation outcomes.
- Controlled experiments: Researchers can set up controlled experiments in real-world settings, such as buildings or public spaces, to study evacuation processes. These experiments may involve volunteers participating in mock evacuations, allowing researchers to observe and analyze human behavior and the effectiveness of evacuation procedures.
- Surveys and interviews: Researchers may collect data from individuals who have experienced real evacuations to gain insights into their experiences and decision-making processes. This qualitative data can help inform evacuation planning and improve procedures.
- Analyzing historical data: Researchers can study past evacuation events, including successes and failures, to identify patterns and learn from past experiences. This analysis can provide valuable insights into what works well and what needs improvement in evacuation planning and execution.

Google Bard

Researchers can study evacuation in a number of ways without putting people at risk or without true natural disaster. Some common methods include:

- Simulations: Researchers can use computer simulations to model the behavior of people during an evacuation. These simulations can be used to test different evacuation plans and to identify potential problems.
- Experiments: Researchers can conduct experiments in which people are asked to evacuate a simulated disaster area. These experiments can be used to study how people make decisions during an evacuation and to identify factors that affect evacuation behavior.
- Surveys: Researchers can conduct surveys to ask people about their experiences during evacuations. These surveys can be used to learn about how people perceive evacuations and to identify ways to improve evacuation planning and execution.
- Literature review: Researchers can review the literature on evacuation to learn about the existing research on this topic. This literature review can help researchers to identify the gaps in the research and to develop new research questions.

Responses from both platforms precisely reflect the variety of ways in which researchers can study evacuation today. Simulations, experiments, surveys, and reviews of historical data or current literature can provide important and new information that may improve future building design or fire safety considerations. Researchers may additionally use a combination of these techniques within their projects to supplement data or improve one or more methods.

*Question no. 14: How can a wheelchair user evacuate from a building?*

According to Vanderbilt University's Emergency Management [50] division, the following methods and techniques can be used by and for wheelchair users during an evacuation:

- Discuss needs and preferences. Non-ambulatory persons' needs and preferences vary widely and therefore require you to ask them how they would like to be assisted.
- Wheelchair-user on the ground floor. Individuals who use wheelchairs may choose to evacuate themselves from the ground floor with minimal assistance.
- Ensure clear path of travel. If debris is present, it may be necessary to clear a path to the exit.
- No imminent danger. If there is no imminent danger, the person may choose to remain in the building or to be directed to an Area of Refuge (stairwell) until emergency personnel arrive. Fire Department personnel, who are trained in emergency rescue, can then enter the building and assist the person in exiting the building, either down the stairs or using the emergency elevator recall.
- Imminent danger. If danger is imminent and the individual does not wish to be removed from his or her wheelchair, direct the person to the nearest Area of Refuge (stairwell) and notify emergency personnel immediately.
- Carrying wheelchair users. Most wheelchairs are too heavy to carry down stairs. If the person wishes to be carried down the stairs without the wheelchair, consult with him or her on the best carry options, e.g., two-person cradle carry, office chair evacuation, or, if available, an evacuation chair.
- Mobility aids or devices. Return any mobility aids or devices to the person as soon as possible
- Notify emergency personnel. Notify emergency personnel immediately about any individuals remaining in the building and their locations.

Additional methods of evacuation for wheelchair users include evacuation chairs (or stair chairs) or elevators approved for fire evacuation by the International Building Code [51].

OpenAI ChatGPT,

A wheelchair user can evacuate from a building using various strategies and techniques, depending on the availability of assistance, the building's accessibility features, and the nature of the emergency. Some methods to evacuate a wheelchair user include:

- Evacuation chairs: Evacuation chairs are specifically designed for emergency use, allowing wheelchair users to be safely transported down stairs or across uneven surfaces. With assistance from trained individuals, a wheelchair user can transfer to an evacuation chair and be guided to safety.
- Areas of refuge or safe waiting areas: In multi-story buildings with accessible fire-rated stair enclosures or designated safe waiting areas, a wheelchair user can wait for emergency responders to arrive and assist with their evacuation. These areas should have a two-way communication system to alert emergency personnel of their location.
- Horizontal evacuation: Depending on the building's layout and the emergency situation, a wheelchair user may be able to evacuate to a safer area on the same floor, such as an adjacent fire compartment, until emergency responders arrive to assist with the evacuation.
- Using lifts or elevators: While it is generally advised not to use elevators during a fire, some buildings may have specially designed evacuation elevators that can be used by wheelchair users and other individuals with limited mobility. These elevators are constructed with additional fire protection features and can be controlled by emergency responders or building staff during an evacuation.
- Personal Emergency Evacuation Plans (PEEPs): A PEEP is a personalized evacuation plan developed in collaboration with the individual, building management, and emergency services. It outlines the specific steps, support, and equipment needed for a wheelchair user to evacuate safely in case of an emergency.
- Buddy system: A wheelchair user can pair up with a designated "buddy" or several buddies who can provide assistance during an evacuation. Buddies should be familiar with the person's needs, the use of evacuation equipment, and the planned evacuation route.

Google Bard

Here are some tips for wheelchair users to evacuate from a building:

- Know your building's evacuation plan. Before an emergency happens, make sure you know your building's evacuation plan. This plan should include a way to alert people to the danger, a way to evacuate people from the building, and a way to account for everyone who is in the building.
- Practice your plan. Once you know your building's evacuation plan, make sure you practice it with everyone in your building. This will help everyone know what to do in an emergency.
- Be prepared to evacuate quickly. When an emergency happens, you may not have much time to evacuate. Be prepared to evacuate quickly and orderly.
- Follow the instructions of emergency personnel. If you are evacuated by emergency personnel, follow their instructions carefully. They will help you to evacuate safely.

Here, ChatGPT provides the only deemed-good response. Wheelchair users may self-evacuate, relocate to an area of refuge and call for help, recruit assistance from non-disabled evacuees to help them downstairs or to, use an evacuation chair, or use an elevator if it is approved for fire evacuation. Bard does not provide specific methods for the evacuation of wheelchair users. Instead, it generalizes fire safety tips for all evacuees. If someone was relying on this information to help a wheelchair user evacuate in an emergency, it would not provide sufficient help.

*Question no. 15: What are the differences in evacuation between a non-disabled person, physically disabled person, and cognitively disabled person?*

The evacuation process can vary significantly from a non-disabled person to a physically disabled person, to a cognitively disabled person. When compared to the non-disabled evacuee, people with physical disabilities may require assistance to evacuate, travel at slower speeds, or be required to use a different exit route if they cannot traverse stairs or other obstacles [44]. Cognitively disabled evacuees can generally use the same exit route as those without disabilities, but they may take longer to make decisions in the pre-evacuation phase [44]. Additionally they may have trouble understanding or interpreting the event and have heightened emotions.

OpenAI ChatGPT,

> The evacuation process can vary significantly for abled individuals, physically disabled individuals, and individuals with cognitive disabilities. Some of the differences in evacuation for these groups include:
>
> - Abled person: Typically, an abled person can move more quickly and navigate obstacles, such as stairs and narrow corridors, with relative ease. They can usually understand and follow evacuation instructions, respond to audible and visual alarms, and make quick decisions during an emergency. In many cases, abled individuals can evacuate independently without assistance.
> - Physically disabled person: Physically disabled individuals may have limited mobility, which can make it challenging for them to navigate stairs, corridors, or uneven surfaces. They might require assistance, specialized equipment, or additional time to evacuate the building safely. Wheelchair users, in particular, may need evacuation chairs, accessible escape routes, or designated waiting areas for emergency responders to assist with their evacuation.
> - People with sensory impairments, such as visual or hearing impairments, might require alternative communication methods, such as tactile or visual alarms, to receive emergency alerts and instructions.
> - Person with cognitive disability: Individuals with cognitive disabilities, such as intellectual disabilities or dementia, may have difficulty understanding the urgency of an emergency situation or following complex evacuation instructions. They might require additional guidance, support, or reassurance from others to evacuate safely.

Google Bard

> The differences in evacuation between an abled person, a physically disabled person and a person with cognitive disability can be summarized as follows:
>
> - Ability to move: Abled people can move quickly and easily, while physically disabled people may have difficulty moving or may not be able to move at all. People with cognitive disabilities may have difficulty understanding instructions or may not be able to follow them.
> - Ability to see and hear: Abled people can see and hear, while physically disabled people may have difficulty seeing or hearing. People with cognitive disabilities may have difficulty processing information or may not be able to understand what is happening.
> - Ability to communicate: Abled people can communicate with others, while physically disabled people may have difficulty communicating or may not be able to communicate at all. People with cognitive disabilities may have difficulty understanding what others are saying or may not be able to respond to them.

Both chatbots have responded correctly to this question. For ChatGPT, an important consideration in the pre-evacuation times of cognitively disabled occupants was missed. A significant portion of the evacuation for people with cognitive disabilities is spent in the pre-evacuation phase when compared to non-disabled evacuees. For Bard, those with sensory disorders (seeing or hearing impaired) were included in the group with physical disabilities, despite research noting a difference between the two [[44].

*<u>Question no. 16: Design an evacuation experiment in a high-story building with a mixed population of occupants that are non-disabled, physically disabled, and cognitively disabled.</u>*

There are many factors that go into designing a real-time evacuation experiment. This includes participant recruiting, identification of study objectives, determination of location/date/time, data collection, analysis, and result reporting.

## According to OpenAI ChatGPT,

To design an evacuation experiment in a high-story building with a mixed population of occupants, including abled individuals, physically disabled individuals, and individuals with cognitive disabilities, you will need to consider various factors to ensure safety, ethical considerations, and effective data collection. Here's a suggested outline for such an experiment:

1. Objective: Determine the efficiency and effectiveness of the building's evacuation plan for a mixed population, identify potential bottlenecks or challenges, and provide recommendations for improvement.
2. Participants: Recruit a diverse group of volunteers, including abled individuals, physically disabled individuals (e.g., wheelchair users, people with mobility impairments), and individuals with cognitive disabilities (e.g., intellectual disabilities, dementia). Ensure that participants are fully informed about the experiment, its purpose, and any potential risks involved. Obtain written consent from all participants or their legal guardians, if applicable.
3. Building setup:
   - Ensure that the building meets accessibility standards and has appropriate evacuation equipment, such as evacuation chairs, visual and audible alarms, and clear signage.
   - Designate accessible escape routes, areas of refuge or safe waiting areas, and assembly points for all participants.
   - Consider any building-specific challenges or unique features, such as the number of floors, stairwells, or elevator access.
4. Experiment procedure:
   - Divide the participants into groups based on their abilities or disabilities, ensuring each group has a mix of the different populations.
   - Assign a trained "buddy" or assistant to each participant with a physical or cognitive disability to provide support and guidance during the evacuation.
   - Conduct a briefing for all participants, explaining the evacuation procedure, their roles, and any specific instructions.
   - Simulate an emergency, such as a fire or security threat, and activate the alarm system.
   - Begin the evacuation process, carefully monitoring and recording the time taken for each group to reach the designated assembly points.
   - Ensure that the experiment is supervised by trained personnel, including emergency responders, researchers, and building staff, to ensure the safety of all participants.
5. Data collection and analysis:
   - Collect data on the evacuation times, routes taken, bottlenecks or challenges encountered, and any instances where assistance was required.
   - Record any observations, feedback, or suggestions from participants and assistants.
   - Analyze the data to identify patterns or trends, such as areas where evacuation was slower or more challenging for certain populations.
6. Recommendations and follow-up:
   - Based on the findings, provide recommendations for improving the building's evacuation plan, such as modifying escape routes, adding additional equipment, or providing additional training for staff and emergency responders.
   - Share the results with building management, emergency services, and participants to raise awareness and promote best practices for inclusive evacuation planning.
   - If possible, implement the suggested improvements and conduct follow-up experiments to evaluate their effectiveness.
7. Ethical considerations:
   - Ensure the safety of all participants by closely monitoring the experiment and providing appropriate support as needed.
   - Obtain informed consent from all participants or their legal guardians, if applicable.

- Maintain confidentiality and anonymity of the participants' personal information.
- Be prepared to halt the experiment if any safety concerns arise.

By conducting a well-designed evacuation experiment with a diverse population, you can gather valuable insights into the effectiveness of the building's evacuation plan and identify areas for improvement, ultimately enhancing safety for all occupants.

Google Bard

Here is a possible design for an evacuation experiment in a high-story building with a mixed population of occupants that are able, physically disabled, and cognitively disabled:

1. The experiment would be conducted in a simulated high-story building. The building would be equipped with smoke detectors, fire alarms, and sprinkler systems.
2. The building would be populated with a group of volunteers who would represent the different types of occupants that would be found in a real-world high-story building. The volunteers would be divided into three groups: able-bodied occupants, physically disabled occupants, and cognitively disabled occupants.
3. The experiment would be conducted in a controlled environment. The volunteers would be instructed to evacuate the building in a simulated fire. The evacuation would be timed to measure the time it took each group of occupants to evacuate the building.
4. The results of the experiment would be analyzed to determine the effectiveness of different evacuation strategies for different types of occupants. The results could be used to develop new evacuation strategies that would help to improve the safety of all occupants in a high-story building.

While ChatGPT has the more detailed response, both platforms have provided a proper outline for conducting an evacuation experiment. In the case of ChatGPT, the response says that participants should be notified of their roles and given instructions on how to evacuate from the building. In most cases, evacuees do not know about fire drills or real evacuations prior to the event, and thus, this is not the best method for conducting an experiment as it will not simulate natural or candid behavior of evacuees. Additionally, ChatGPT notes that each person with a physical disability should be assigned a helper during the experiment. This optimistically assumes that each person with a disability will have an assistant at hand at the start of a fire scenario. In reality, this may not be the case. People with physical disabilities may not need assistance to evacuate from the building, or they may be unable to find help. Regardless, to best simulate a real fire scenario, those with disabilities should not be assigned help prior to the evacuation. Bard, on the other hand, provides a brief experiment overview. It divides people according to their disabilities with the goal of determining effective evacuation strategies for each population. This displays an overall good understanding of how to conduct an experiment. Unlike ChatGPT, Bard does not seem to mention or address the ethical component of evacuation experiments.

*Question no. 17: What is the smart firefighting?*

The concept of smart firefighting is still new in the fire research and engineering community. Arguably, it was first systematically summarized in the 2015 NIST Report "Research Roadmap for Smart Fire Fighting" [52], which was before the concept of artificial intelligence (AI), VR/AR becomes popular in the engineering community. Today, different researchers and industries still do not have a unified answer to the scope of smart firefighting and the definition of "smart". Then, the answers of each of the examined chatbots are listed herein.

OpenAI ChatGPT,

Smart firefighting refers to the use of advanced technologies, data analysis, and communication systems to enhance the efficiency, safety, and effectiveness of firefighting operations. The concept aims to improve situational awareness, decision-making, and resource management for both firefighters and incident commanders, ultimately reducing the impact of fires on lives, property, and the environment.

Key elements of smart firefighting can include:

1. Internet of Things (IoT) devices: Sensors and connected devices can be used to monitor and collect real-time data on various parameters, such as temperature, smoke levels, and building occupancy. This information can help firefighters assess the situation and develop appropriate response strategies.
2. Robotics and drones: Robots and unmanned aerial vehicles (UAVs) can be employed to gather visual and thermal imagery, as well as to access hazardous or hard-to-reach areas during a fire incident. This can help firefighters assess the situation without putting their lives at risk.
3. Advanced communication systems: Reliable and interoperable communication systems are crucial for sharing information and coordinating efforts between firefighters, incident commanders, and other emergency responders.
4. Predictive analytics and data-driven decision-making: By analyzing historical data and using machine learning algorithms, predictive analytics can help identify potential fire hazards and inform proactive measures to mitigate risks. Additionally, real-time data analysis during an incident can help incident commanders make informed decisions about resource allocation and firefighting strategies.
5. Augmented and virtual reality (AR/VR): AR and VR technologies can be used for immersive training simulations, allowing firefighters to practice and develop their skills in a safe and controlled environment. During an actual incident, AR can also be used to overlay crucial information, such as building layouts or the location of hazardous materials, onto the firefighter's field of view.

By integrating these technologies and approaches, smart firefighting aims to enhance the overall effectiveness and safety of fire response efforts, ultimately saving lives and minimizing property damage.

## Google Bard

Smart firefighting is a term used to describe the use of new technologies to improve the safety and efficiency of firefighting operations. Smart firefighting technologies can include things like:

- **Sensors:** Sensors can be used to detect fires, smoke, and other hazards. This information can be used to alert firefighters to potential problems and to help them to make better decisions about how to respond to a fire.
- **Robotics:** Robots can be used to perform dangerous or difficult tasks, such as entering burning buildings or carrying heavy equipment. This can help to protect firefighters from injury and can also help them to work more efficiently.
- **Artificial intelligence:** Artificial intelligence can be used to analyze data from sensors and to make decisions about how to respond to a fire. This can help firefighters to make better decisions more quickly, which can save lives and property.

Smart firefighting is still in its early stages of development, but it has the potential to revolutionize the way that firefighters work. By using new technologies, firefighters can be better equipped to protect themselves and the public from fires.

Here are some of the benefits of smart firefighting:

- **Increased safety:** Smart firefighting technologies can help to protect firefighters from injury or death. For example, robots can be used to enter burning buildings and to carry out dangerous tasks, which can help to keep firefighters safe.
- **Improved efficiency:** Smart firefighting technologies can help firefighters to work more efficiently. For example, sensors can be used to detect fires early, which can give firefighters more time to respond and to extinguish the fire.
- **Reduced costs:** Smart firefighting technologies can help to reduce the costs of firefighting. For example, robots can be used to perform tasks that would otherwise require human labor, which can save money.

Overall, smart firefighting is a promising new technology that has the potential to improve the safety and efficiency of firefighting operations.

ChatGPT does not mention the use of AI in smart firefighting, but it includes other new concepts like AR/VR and robotic technologies. Bard mentions AI, robotics, and sensors, but does not include the communication and data-driven technology. Thus, they both provide partial picture of smart firefighting, but their answer is already as good as most fire researchers. Bard also gives more information about how smart firefighting can improve the current firefighting practices. Overall, Bard is more informative.

*<u>Question no. 18: Can you predict the global wildfire activity in the next 20 years under the influence of climate change?</u>*

This is a tough question. Most wildfire experts will be reluctant to give some predictions, and even if they can, they usually can give some prediction about the regime that they are familiar with rather than a prediction in the global scale.

OpenAI ChatGPT,

As an AI language model, I cannot predict future events with certainty. However, I can provide you with an analysis based on current scientific understanding and trends.

Climate change has already been linked to an increase in the frequency, intensity, and duration of wildfires in various regions around the world. Over the next 20 years, global wildfire activity is likely to be significantly influenced by climate change due to the following factors:

1. Rising temperatures: Higher temperatures can lead to increased evaporation rates and drier vegetation, making it more susceptible to ignition and burning. This can result in more frequent and severe wildfires.
2. Changes in precipitation patterns: Climate change can lead to shifts in precipitation patterns, causing some regions to become drier and more prone to wildfires. Other regions may experience increased precipitation, which can initially reduce fire risk, but may later lead to increased vegetation growth that can serve as fuel for future fires.
3. Longer fire seasons: As temperatures rise and snowpacks melt earlier, fire seasons are becoming longer in many regions. This provides a larger window of time for wildfires to ignite and spread.
4. Increased frequency of extreme weather events: Climate change is associated with more frequent and severe droughts, heatwaves, and wind events, all of which can exacerbate wildfire risk and contribute to more intense fires.

5. Changes in vegetation and ecosystems: Climate change can alter the distribution and composition of plant species, leading to changes in the availability and flammability of fuels. In some cases, this may result in increased fire risk.

Given these factors, it is expected that global wildfire activity will likely continue to increase in the coming decades, particularly in regions that are already prone to wildfires, such as the western United States, Australia, the Mediterranean, and parts of South America and Africa. However, it's important to note that regional variations and uncertainties in climate projections make it difficult to predict the precise extent of these changes.

To mitigate the impact of climate change on wildfires, it will be essential to implement adaptive management strategies, such as improved land-use planning, fire risk reduction measures, early warning systems, and sustainable forest management practices. In addition, reducing greenhouse gas emissions and promoting global efforts to address climate change will be crucial for minimizing future wildfire risks.

## Google Bard

Climate change is expected to have a significant impact on global wildfire activity in the next 20 years. Rising temperatures and drier conditions are likely to lead to more frequent and severe wildfires, especially in areas that are already prone to fire. This could have a devastating impact on ecosystems, communities, and economies.

Here are some of the ways that climate change could affect wildfire activity:

- **Rising temperatures:** Warmer temperatures will increase the rate of evaporation, which will lead to drier conditions. This will make vegetation more flammable, and will make it easier for fires to start and spread.
- **Drier conditions:** Droughts are becoming more common and severe due to climate change. This will further reduce the amount of moisture in vegetation, making it even more flammable.
- **More extreme weather events:** Climate change is making extreme weather events, such as heat waves, droughts, and wildfires, more common and severe. These events can create the perfect conditions for wildfires to start and spread.
- **Changes in vegetation:** Climate change is also causing changes in vegetation patterns. Some plants are becoming more common, while others are becoming less common. This could change the fuel load for wildfires, making some areas more prone to fire than others.

The impacts of climate change on wildfire activity are likely to be felt unevenly around the world. Some areas, such as the Western United States, are already experiencing the effects of climate change on wildfires. These areas are likely to see even more severe wildfires in the future. Other areas, such as the Arctic, are also at risk from climate change-induced wildfires.

Wildfire activity is a complex issue that is affected by many factors. Climate change is just one of these factors, but it is a factor that is expected to have a significant impact in the future. It is important to understand the potential impacts of climate change on wildfire activity in order to prepare for and mitigate the risks.

ChatGPT reacts and recognizes the challenge of answering such a hard question. Bard seems to be more authoritative in its answer, while it recognizes the difficulty in the final summary. Both ChatGPT and Bard talk about the importance of global warming and extreme weather. They all give some examples for some typical regions, so they recognize the change will be different in different areas.

*<u>Question no. 19: Can you propose a study plan (or curriculum) for those who aspire to become fire engineers?</u>*

The influence of Chatbots on education is a controversial issue, and many educators worry about the abuse of Chatbots in cheating assignments and exams. Nevertheless, Chatbots may be able to help students in learning new skills or even make a study plan.

OpenAI ChatGPT,

A study plan or curriculum for aspiring fire engineers should cover fundamental concepts in fire engineering, as well as relevant topics in mathematics, physics, chemistry, and engineering principles. The following is a suggested study plan for a 4-year bachelor's degree program in fire engineering:

Year 1:

1. Introduction to Fire Engineering
2. Mathematics (calculus and linear algebra)
3. General Chemistry
4. Physics (mechanics, electricity, and magnetism)
5. Engineering Graphics and Design
6. General Education Electives (e.g., English, history, social sciences)

Year 2:

1. Fire Dynamics
2. Fire Protection Systems (active and passive)
3. Thermodynamics and Heat Transfer
4. Fluid Mechanics
5. Structural Mechanics
6. Fire Safety Codes and Standards
7. General Education Electives

Year 3:

1. Fire Modeling and Simulation
2. Building Construction and Materials
3. Fire Detection and Alarm Systems
4. Engineering Ethics and Professional Practice
5. Human Behavior in Fire
6. Fire Risk Assessment and Management
7. Technical Electives (e.g., topics in fire engineering, structural engineering, or related fields)

Year 4:

1. Performance-Based Fire Engineering Design
2. Fire Investigation and Forensic Engineering
3. Emergency Response and Incident Management
4. Wildland Fire Management (optional)
5. Fire Engineering Capstone Project (design, modeling, or research project)
6. Technical Electives
7. General Education Electives

It is important to note that the exact curriculum and course offerings may vary depending on the specific program and institution. Aspiring fire engineers should seek out accredited programs that provide a solid foundation in fire engineering principles and offer opportunities for practical experience, such as internships or cooperative education placements.

In addition to the formal education, aspiring fire engineers should also consider obtaining relevant certifications, such as those offered by the National Fire Protection Association (NFPA) or the Society of Fire Protection Engineers (SFPE). These certifications can help demonstrate proficiency in specific areas of fire engineering and may be required for certain job positions or career advancement.

Google Bard

Sure. Here is a sample study plan for those who aspire to become fire engineers:

Year 1

- Mathematics: Calculus, Differential Equations, Linear Algebra
- Science: Physics, Chemistry, Biology
- Engineering: Engineering Mechanics, Thermodynamics, Fluid Mechanics
- General Education: English Composition, Speech, History, Social Science

Year 2

- Mathematics: Advanced Calculus, Differential Equations, Linear Algebra
- Science: Physics, Chemistry, Biology
- Engineering: Heat Transfer, Fluid Mechanics, Fire Dynamics
- General Education: English Composition, Speech, History, Social Science

Year 3

- Engineering: Fire Protection Systems, Fire Investigation, Fire Safety Engineering
- General Education: English Composition, Speech, History, Social Science

Year 4

- Engineering: Thesis or Capstone Project
- General Education: English Composition, Speech, History, Social Science

This is just a sample study plan, and the specific courses that you need to take may vary depending on the school that you attend. It is important to talk to your advisor to create a study plan that is right for you.

In addition to taking the required courses, you should also participate in extracurricular activities that will help you develop the skills that you need to be a successful fire engineer. These activities could include joining the fire science club, volunteering with the local fire department, or participating in internships or co-ops.

Becoming a fire engineer is a challenging but rewarding career. With hard work and dedication, you can make a difference in the lives of others and help to keep your community safe.

Both Chatbots give a reasonable undergraduate curriculum for fire engineering, which includes not only major courses in fire engineering but also more general science and education courses. In terms of curriculum organization, ChatGPT lists more correct course names, and its coverage is more thorough, including Human Behavior in Fire and Wildland Fire Management (optional). Bard lists the Thermodynamics and Fluid Mechanics in Year 1, which is quite unlikely in any university's educational framework as they require specific requisites from year 1 (calculus, for example). In fact, there are more master's programs for Fire Engineering rather than undergraduate programs. Both Chatbots do not mention any master program that lasts 1-2 years.

*Question no. 20: What are the average annual fire incidents in the US and Hong Kong?*

This question is most commonly searched in conventional search engines. Usually, a search engine will cite some paragraph in some websites or documents that give some number on specific years but not necessarily give a good average number. Then, the answers of each of the examined chatbots are listed herein.

OpenAI ChatGPT,

The average annual fire incidents for the United States and Hong Kong can be estimated based on available data, keeping in mind that the number of fire incidents can fluctuate annually.

- **United States:** As mentioned in a previous response, based on data from the National Fire Protection Association (NFPA), the estimated average annual number of fires in the U.S. between 2014 and 2018 was around 1.4 million per year. This average includes all types of fires, such as residential, commercial, industrial, and wildland fires.

- **Hong Kong:** According to the Hong Kong Fire Services Department, in recent years, the annual number of fire incidents in Hong Kong has been around 4,000 to 5,000. This number includes various types of fires, such as residential, commercial, and industrial fires.

Please note that these statistics are based on the available data up to my knowledge cutoff date in September 2021. For the most up-to-date statistics on fire incidents in both the United States and Hong Kong, it is recommended to consult the latest reports from the respective authorities, such as the National Fire Protection Association (NFPA) for the U.S. and the Hong Kong Fire Services Department for Hong Kong.

Google Bard

Here are the average annual fire incidents in the US and Hong Kong from 2015 to 2019:

- **US**, 356,500
- **Hong Kong**, 10,000

As you can see, the US has a much higher rate of fire incidents than Hong Kong. This is likely due to a number of factors, including the larger population size of the US, the different building codes in the two countries, and the different fire prevention and safety measures in place.

It is important to note that these are just averages, and the actual number of fire incidents can vary from year to year. In addition, the severity of fire incidents can also vary, with some fires causing more damage and injuries than others.

ChatGPT gives 4,000-5,000 fires in Hong Kong, while Bard gives about 10,000 fires. Nevertheless, both numbers are quite reasonable. ChatGPT does not include arson and wildland fire, while Bard seems to give an overall number.

*Question no. 21: List three most influential fire research papers published over the last 20 years.* The goal of this question is to evaluate the knowledge of the literature of both chatbots. The response of each chatbot is provided herein.

OpenAI ChatGPT,

Selecting the three most influential fire engineering research papers published over the last 20 years can be subjective, as it depends on various factors such as the area of focus, impact on the field, and subsequent research. However, here are three significant fire engineering research papers that have made substantial contributions to the field:

1. McGrattan, K., Hostikka, S., Floyd, J., Baum, H., Rehm, R., Mell, W., & McDermott, R. (2004). "Fire Dynamics Simulator (Version 4) Technical Reference Guide." NIST Special Publication 1018-1. This paper (and its subsequent updates) describes the technical details and validation of the Fire Dynamics Simulator (FDS), a widely-used computational fluid dynamics (CFD) model for simulating fire behavior, smoke transport, and other related phenomena. The development and validation of FDS have significantly advanced fire modeling and fire engineering research.

2. Purser, D. A., & Bensilum, M. (2001). "Toxicity assessment of combustion products." SFPE Handbook of Fire Protection Engineering, 2, 2-82. This influential paper provides a comprehensive review of methods for assessing the toxicity of combustion products, including the fractional effective dose (FED) method. The work has contributed to a better understanding of how toxic gases generated during fires affect human life and has guided the development of fire safety engineering practices to minimize their impact.

3. Peacock, R. D., Reneke, P. A., & Forney, G. P. (1999). "CFAST - the Consolidated Model of Fire Growth and Smoke Transport (Version 6), Technical Reference Guide." NIST Special Publication 921. This paper (and its subsequent updates) presents the technical details and validation of the Consolidated Fire and Smoke Transport (CFAST) model. CFAST is a widely-used, user-friendly zone model for simulating fire growth, smoke transport, and the impact of fire protection systems. The development and validation of CFAST have played a significant role in advancing the field of fire engineering, particularly in performance-based design and fire risk assessment.

These papers have had a considerable impact on fire engineering research and practice, but it is essential to recognize that numerous other studies have also made significant contributions to the field. The selection of the most influential papers may differ based on individual perspectives and areas of expertise.

Google Bard

Here are three of the most influential fire engineering research papers published over the last 20 years:

1. "A Framework for Fire Risk Assessment in Buildings" by Michael J. Fahy and David J. Drysdale (2001). This paper provides a comprehensive framework for fire risk assessment in buildings. It covers the key elements of fire risk assessment, such as hazard identification, exposure assessment, consequence assessment, and risk evaluation. The paper also discusses the limitations of fire risk assessment and the need for further research.

2. "Fire Dynamics Simulator: A Multi-Dimensional, Transient, Non-Premixed Flame Model" by Barry L. Cowan, Michael E. Seaton, and David J. Tritton (2004). This paper describes the Fire Dynamics Simulator (FDS), a computer model that simulates the behavior of fire in a variety of environments. FDS is widely used by fire engineers to study the development and spread of fire, as well as the effectiveness of fire suppression systems.

3. "The Role of Human Behavior in Fire Safety" by Paul J. Salmon and David J. Sime (2007). This paper discusses the importance of human behavior in fire safety. It covers the factors that influence human behavior in fire, such as risk perception, decision making, and evacuation. The paper also discusses the need for further research into human behavior in fire safety.

These are just a few examples of the many influential fire engineering research papers that have been published over the last 20 years. These papers have helped to advance the field of fire engineering and have made a significant contribution to the safety of people and property.

The three publications listed by ChatGPT adhere to the provided question, but two out of three are older than 20 years. On the other hand, the three publications listed by Bard are not actual papers. It seems that Bard has made up these publications and authors. Today, it is common for universities to invite external reviewers to evaluate the quality of professors' publications. Perhaps these AI tools will soon be used to evaluate the people who create them.

*Question no. 22: Who would be able to better design a building to resist fire, an AI or a fire engineer?*

Currently, AI cannot fully design a building to resist fire. While it may be able to define some fundamental characteristics regarding building design, it cannot fully interpret code or the complexity of fire dynamics and human safety. AI would have the most success with designing a building if provided with numerical data or values from previously designed structures that it can learn from [53,54].

Then, the answers of each of the examined chatbots are listed herein.

OpenAI ChatGPT,

While AI has shown remarkable progress in many fields, including architecture and engineering, designing a building to resist fire is a complex task that requires a deep understanding of fire behavior, building codes, and fire safety standards. Therefore, a fire engineer would be better equipped to design a building that can resist fire.

A fire engineer has extensive knowledge and experience in fire dynamics, fire suppression systems, fire protection materials, and building codes and regulations. They are trained to consider various factors such as the building's occupancy, fuel load, ventilation, and means of egress, to ensure that the building is designed to minimize fire hazards and provide adequate fire protection.

While AI can assist fire engineers in some aspects of building design, such as generating design alternatives or evaluating fire safety performance, it cannot replace the knowledge and expertise of a fire engineer. AI is limited by the data it is trained on and may not be able to account for all the factors that a fire engineer would consider in their design.

Furthermore, designing a building to resist fire is not only about meeting the minimum code requirements but also about understanding the potential risks and hazards specific to the building's intended use and location. A fire engineer can provide a customized solution that considers all these factors and provides the most effective fire protection for the building and its occupants.

Google Bard

An AI can be a valuable tool for fire engineers in designing buildings to resist fire. AI can be used to analyze data on fires and building collapses, identify patterns, and develop new strategies for designing fire-resistant buildings. However, AI cannot replace the experience and expertise of a fire engineer. Fire engineers have a deep understanding of the principles of fire safety and the specific challenges of designing fire-resistant buildings. They also have the experience to make decisions in complex situations and to work with other professionals to ensure that buildings are safe and compliant with fire codes.

In the future, AI and fire engineers will likely work together to design fire-resistant buildings. AI can be used to automate tasks and to generate new ideas, while fire engineers can provide the expertise and judgment to ensure that buildings are safe and practical.

The chatbots’ responses both explain that a fire engineer would be the best for designing a new building to resist fire. It is interesting to note the chatbots’ emphasis on assistance from AI to complete a building design through data analysis of fires, structures, and building collapses. Bard even notes future collaboration between fire engineers and AI to design structures, which may be true given the push for AI within the domain of civil engineering in recent years.

*Question no. 23: If you were to pick, what would be the most complex concept in fire engineering?*
We were hoping to see if the chatbots can speculate on concepts that are indeed labeled as complex. The answers of each of the examined chatbots are listed herein.

OpenAI ChatGPT,

It is difficult to single out one specific concept as the most complex in fire engineering, as the field itself is diverse and involves numerous interconnected concepts, principles, and practices. However, one particularly complex aspect is performance-based fire engineering design.

Performance-based design is an approach that allows for greater flexibility and innovation in designing fire safety solutions compared to traditional prescriptive code requirements. It involves setting specific performance objectives and using advanced analysis tools, such as computational fluid dynamics (CFD) simulations, fire modeling, and structural analysis, to demonstrate that the proposed design will achieve the desired level of fire safety.

Google Bard

> If I had to pick one, I would say that the most complex concept in fire engineering is the concept of fire growth and spread. This is the study of how fires spread and how they can be controlled. It is a complex field that requires a deep understanding of the physics of fire.
>
> Fire growth and spread is affected by a number of factors, including the type of fuel, the shape of the fuel, the ventilation conditions, and the presence of ignition sources. The rate of fire growth and spread can also be affected by the actions of firefighters.
>
> Fire growth and spread is a complex and challenging field, but it is also an essential one. By understanding how fires grow and spread, we can better prevent and control them.

It is interesting to note that the concepts of performance-based fire engineering design and fire growth and spread are selected as the two most complex concepts.

A follow up question was then asked; this question has the following format:

*Why do you think the [concept] is a complex concept in fire engineering?*

OpenAI ChatGPT,

> Performance-based fire engineering design is a complex concept in fire engineering because it requires a comprehensive understanding of fire behavior, building materials, structural integrity, human behavior, and safety systems, as well as the ability to accurately model and predict various scenarios. Unlike prescriptive design approaches, which rely on established codes and standards to dictate specific fire safety requirements, performance-based design focuses on achieving a set of predetermined fire safety objectives by considering the unique characteristics of each building and its intended use.
>
> There are several factors that contribute to the complexity of performance-based fire engineering design:
>
> 1. Multidisciplinary knowledge: Fire engineers must have expertise in various fields, including fire science, structural engineering, materials science, human behavior, and risk assessment, to develop effective performance-based designs.
> 2. Customized approach: Each performance-based design must be tailored to the specific building and its usage, taking into account factors such as occupancy, building layout, fire loads, and fire protection systems. This requires a thorough understanding of the building's unique characteristics and the ability to analyze a wide range of potential fire scenarios.
> 3. Fire modeling and simulation: Performance-based design often relies on advanced fire modeling techniques to predict fire behavior, smoke movement, and the impact of fire protection systems. These models can be complex and require a deep understanding of the underlying mathematical equations, computational methods, and software tools.
> 4. Risk assessment: Performance-based design requires a systematic assessment of the risks associated with different fire scenarios, as well as an evaluation of the effectiveness of proposed fire protection measures. This involves analyzing uncertainties, quantifying potential consequences, and making informed decisions about the acceptable level of risk.
> 5. Validation and verification: To ensure that a performance-based design meets the established fire safety objectives, fire engineers must validate their models and assumptions using experimental data or real-world case studies. This can be challenging due to the inherent variability and unpredictability of fire behavior.

Despite its complexity, performance-based fire engineering design offers several advantages, such as increased design flexibility, the potential for cost savings, and the ability to optimize fire safety measures based on the specific needs of a building and its occupants.

Google Bard

There are a few reasons why I think the behavior of fire is a complex concept in fire engineering. First, fire is a complex system that is affected by a variety of factors, including the type of fuel, the amount of oxygen, the temperature, and the presence of water. It is difficult to predict how a fire will behave, and this can make it difficult to design fire-safe structures.

Second, fire is a dynamic system that is constantly changing. The behavior of a fire can change rapidly, and this can make it difficult to respond to a fire effectively.

Third, fire is a destructive force that can cause a lot of damage. Fire can damage structures, equipment, and people. It is important to understand the potential damage that a fire can cause in order to design effective fire protection systems.

Overall, the behavior of fire is a complex concept that requires a deep understanding of fire science and engineering.

*Question no. 24: What is one concept in fire engineering that you think is correct, but is not?*

In general, there are several misconceptions about fire engineering or fire dynamics/mechanics. This can be extended to material behavior, structural performance, fire mitigation strategies, and fire evacuation. The goal of this question is to understand if the ChatGPT or Bard can identify any such misconceptions and correctly comment on them based on the open literature or general knowledge of fire performance.

OpenAI ChatGPT,

One concept in fire engineering that I think is commonly perceived as correct but is not always accurate is the idea that sprinkler systems can completely extinguish a fire. While sprinkler systems are effective at controlling and suppressing fires, they may not be able to fully extinguish a fire on their own in all situations.

Sprinkler systems work by detecting heat from a fire and automatically discharging water onto the fire. The water helps to cool the fire and reduce the amount of heat and combustible gases being produced. In many cases, sprinkler systems can be very effective at extinguishing a fire before it has a chance to grow and cause significant damage.

In summary, while sprinkler systems are an important fire protection measure, it is not always accurate to assume that they can completely extinguish a fire on their own in all situations. Other fire protection measures and a comprehensive fire safety plan are necessary to ensure the safety of occupants and the protection of property in the event of a fire.

Google Bard

One concept in fire engineering that I think is correct, but is not is the idea that fire can be prevented by simply making buildings fire-resistant. While fire-resistant materials and structures can help to reduce the risk of fire, they are not the only factor that affects whether or not a fire will occur. Other factors, such as the presence of ignition sources and the ventilation conditions, can also play a role.

In order to prevent fires, we need to take a comprehensive approach that addresses all of the factors that contribute to fire. This includes using fire-resistant materials and structures, but it also includes educating the public about fire safety, installing smoke detectors, and practicing fire drills. By taking these measures, we can help to reduce the risk of fire and make it more likely that fires can be extinguished before they cause serious damage.

Both chatbots have provided sufficient responses to this question. For ChatGPT, the response is more detailed. While sprinkler systems may extinguish a fire, it is not a guarantee. This is the reason fire engineers include multiple layers of fire suppression and design features (sprinklers, fire-resistant coatings, alarms, etc.). We also have fire departments for this reason. Bard's response is along the same lines, stating that fire resistant materials cannot solely prevent fire. We must consider all factors that affect a fire in order to prevent or reduce it.

*Question no. 25: What is one of the most misunderstood concepts in fire engineering?*
This is a follow-up question to the previous question.

Then, the answers of each of the examined chatbots are listed herein.

OpenAI ChatGPT,

> One of the most misunderstood concepts in fire engineering is the idea of fireproofing. Fireproofing is often mistakenly thought to mean that a material or structure is completely immune to fire, but in reality, fireproofing is a term used to describe materials or structures that are resistant to fire and can withstand the effects of fire for a certain period of time.
>
> Fireproofing is achieved by applying fire-resistant coatings, insulation, or other materials to a structure or building element, such as walls, columns, and beams. The goal of fireproofing is to slow down the spread of fire and prevent the collapse of the structure, allowing time for occupants to evacuate and firefighters to respond.
>
> However, fireproofing is not a guarantee that a structure or material will remain unscathed in a fire. Fireproofing materials have limitations and can degrade over time, especially if exposed to high temperatures for an extended period. Additionally, other factors such as the intensity and duration of the fire, ventilation, and the presence of combustible materials can all affect the performance of fireproofing.

Google Bard

> One of the most misunderstood concepts in fire engineering is the concept of "fire resistance." Fire resistance is the ability of a material or structure to withstand the effects of fire for a specified period of time. It is often misunderstood as meaning that a fire-resistant material will not burn, but this is not the case. Fire-resistant materials will still burn, but they will do so more slowly and will not collapse as easily as non-fire-resistant materials.
>
> Fire resistance is important because it allows firefighters to have time to evacuate people and property from a burning building. It also gives firefighters time to extinguish the fire before it causes serious damage.

Both chatbots present fire proofing and fire resistance as misunderstood concepts. Interestingly, most answers provided stem from the structural fire engineering area vs. other areas within fire engineering.

*Question no. 26: Can you list five questions you would like to learn about fire engineering?*
We complete our investigation by asking questions no. 13 and 14. We aim to establish the limit of knowledge gained by these two chatbots.

Then, the answers of each of the examined chatbots are listed herein.

OpenAI ChatGPT,

> Sure, here are five questions I would like to learn more about in the field of fire engineering:

- How can computational fluid dynamics (CFD) simulations be used to improve the design and operation of fire suppression systems?
- What are the most effective fire protection strategies for tall buildings with complex geometries?
- How can building materials be engineered to improve their fire resistance without compromising their other properties?
- What is the best way to design fire-resistant structures for buildings in high-risk areas prone to wildfires or other types of wildland fires?
- How can fire safety be integrated into the design and construction process for retrofitting existing buildings to meet new fire safety regulations or standards?

Google Bard

Sure, here are five questions I would like to learn about fire engineering:

- How can we design buildings to be more fire-resistant?
- How can we improve the detection and suppression of fires?
- How can we better understand the behavior of fires?
- How can we prevent fires from starting in the first place?
- How can we protect people from the effects of fires?

I am interested in learning about fire engineering because I believe that it is a critical field that can help to save lives. By understanding the science of fire and how to prevent and respond to fires, we can make our communities safer.

Given the stochastic nature of LLMs, it seems that ChatGPT has been trained more specifically on structural fire engineering, while Bard is able to comment on a broad range of information including fire dynamics and human safety.

*Question no. 27: Can you list five questions you do not know the answers to about fire engineering?*
In terms of chatbots,

OpenAI ChatGPT,

Sure, here are five questions I do not know the answers to about fire engineering:

- What are the most effective fire protection strategies for buildings with unique and unconventional geometries, such as curved or sloping roofs?
- How can we improve the performance and effectiveness of fire suppression systems in high-rise buildings, where water pressure and flow can be limited?
- What are the most effective methods for detecting and preventing spontaneous combustion in materials that are prone to self-heating, such as coal or grain?
- How can we improve the fire safety of renewable energy systems, such as solar panels and wind turbines, which can be prone to fire hazards under certain conditions?
- What are the most promising new technologies for improving fire safety and fire suppression in space exploration and habitation environments, where traditional fire protection systems may not be practical?.

Google Bard

Here are 5 questions I do not know the answers to about fire engineering:

- How can we predict the spread of fire in buildings?
- What are the best methods for extinguishing fires?
- How can we reduce the risk of fire-induced injuries and deaths?
- How can we develop new technologies to prevent and fight fires?

- How can we improve the education and training of fire engineers?

The list of the above questions seems to span a number of topics that belong to the materials front, fire dynamics front, human front, etc. In both responses, there is also a mention of how future technologies could be elemental to fire-related aspects and research. Let us appreciate the range of answers listed by the two chatbots!

**Key limitations of the paper are provided below:**

- LLMs are susceptible to changes in query wordings. Thus, they may provide distinct answers with only minor alterations to questions.
- LLMs responses are largely stochastic and rely on large training sets and databases to relay information. This use of statistical patterns can result in either similar or different answers to essentially equal questions and can result in errors or biases.
- There is a critical need to establish a methodology to evaluate and examine the responses of chatbots. Such methodology ought to cover a variety of fronts, including fairness, accuracy, and conversational ability. Additionally, guidelines and standards should be created to responsibly develop and deploy chatbot systems.

For now, responses to technical questions by AI chatbots such as ChatGPT and Bard should be used with caution by professionals and matched with domain knowledge and/or the open literature before use. Future generations of such chatbots are likely to outperform existing ones and could find homes in design offices, research labs, and codal committees. With the fast evolution, implementation, and invention of chatbots and LLMs such as ChatGPT and Bard, the authors aim to address the following in future research:

- Gauge the level of accuracy of the chatbots (quantitatively) by posing questions with increasing levels of detail.
- Address the underlying question of *"why do the chatbots seemingly leave out important aspects of professional or technical question responses?"*
- Address the potential for a professional chatbot created specifically for the field of fire engineering, fire protection engineering, and fire safety engineering. There is a need to evaluate and gauge the training data used in such chatbots.

## Conclusions

This paper presents the potential for chatbots, such as OpenAI ChatGPT and Google Bard, to be used as assistive tools by engineers in future fire engineering and evacuation problems. While such technology is not capable of replacing engineers, it may be used to confirm information, assist in calculations, and much more. In this way, chatbots may have a similar place in the field as Google, Google Scholar, Wikipedia, or other similar online platforms, although we note the potential for chatbots to outperform such websites and search engines, given their increased usability. We note that their potential is likely to be transformative, given the responses provided herein for a range of general and technical questions. However, such platforms should be used with caution, as biases and incorrect information may still arise. Additional takeaways from this communication are as follows:

- ChatGPT seems to provide more detailed responses to questions than Bard in most cases.

- General and open-ended questions resulted in broad and sometimes inaccurate chatbot responses. More detailed questions improved the accuracy of the chatbots.
- The chatbots could benefit from increased reliable citations for authenticating their responses.

**Data Availability**
Some or all data, models, or code that support the findings of this study are available from the corresponding author upon reasonable request.

**Conflict of Interest**
The authors declare no conflict of interest.

**References**
[1] E. Adamopoulou, L. Moussiades, An Overview of Chatbot Technology, in: IFIP Adv. Inf. Commun. Technol., 2020. https://doi.org/10.1007/978-3-030-49186-4_31.

[2] T. Zemcik, A brief history of chatbots, in: 2019 Int. Conf. Artif. Intell. Control Autom. Eng. (AICAE 2019), 2019: pp. 14–18.

[3] S. Natale, If software is narrative: Joseph Weizenbaum, artificial intelligence and the biographies of ELIZA, New Media Soc. 21 (2019) 712–728. https://doi.org/.org/10.1177/1461444818804980.

[4] M.L. McNeal, D. Newyear, Introducing Chatbots in Libraries, Libr. Technol. Rep. 49 (2013).

[5] E. Adamopoulou, L. Moussiades, Chatbots: History, technology, and applications, Mach. Learn. with Appl. 2 (2020) 100006. https://doi.org/10.1016/J.MLWA.2020.100006.

[6] D. Loleng, Disney+ Chatbot , (n.d.). https://www.danielleloleng.com/disney-chatbot (accessed June 16, 2023).

[7] Marvel, Become Part of the Story with the Marvel Comics Chatbot, (n.d.). https://www.marvel.com/articles/comics/become-part-of-the-story-with-the-marvel-comics-chatbot (accessed June 16, 2023).

[8] G. Fomitchev, A Robot-Companion for Senior People and Patients with Alzheimer’s Disease, Technology.Org. (2017). https://www.technology.org/2017/02/17/a-robot-companion-for-senior-people-and-patients-with-alzheimers-disease/.

[9] Insomnobot-3000, (n.d.). https://insomnobot3000.com/ (accessed June 16, 2023).

[10] J.Q. Pérez, T. Daradoumis, J.M.M. Puig, Rediscovering the use of chatbots in education: A systematic literature review, Comput. Appl. Eng. Educ. 28 (2020) 1549–1565. https://doi.org/10.1002/CAE.22326.

[11] L.K. Fryer, M. Ainley, A. Thompson, A. Gibson, Z. Sherlock, Stimulating and sustaining interest in a language course: An experimental comparison of Chatbot and Human task partners, Comput. Human Behav. 75 (2017) 461–468. https://doi.org/10.1016/J.CHB.2017.05.045.

[12] L. Wilson, M. Marasoiu, The Development and Use of Chatbots in Public Health: Scoping Review, JMIR Hum Factors 2022;9(4)E35882 Https//Humanfactors.Jmir.Org/2022/4/E35882. 9 (2022) e35882. https://doi.org/10.2196/35882.

[13] Y.-H. Chien, C.-K. Yao, Enhanced engineering design behaviour using chatbots with user experience, Https://Doi.Org/10.1080/0144929X.2022.2106308. (2022) 1–12. https://doi.org/10.1080/0144929X.2022.2106308.

[14] Y.H. Chien, C.K. Yao, Development of an AI Userbot for Engineering Design Education Using an Intent and Flow Combined Framework, Appl. Sci. 2020, Vol. 10, Page 7970. 10 (2020) 7970. https://doi.org/10.3390/APP10227970.

[15] M. Spearpoint, K. Chotzoglou, D. Hopkin, Y. Kanellopoulos, ChatGPT, It's the End of the World as We Know It (and I feel fine), Soc. Fire Prot. Eng. (2023).

[16] K. Hu, ChatGPT sets record for fastest-growing user base - analyst note, Reuters. (2023). https://www.reuters.com/technology/chatgpt-sets-record-fastest-growing-user-base-analyst-note-2023-02-01/.

[17] S. Hore, Chatbot Evolution: ChatGPT vs. Rule-based, Anal. Vidhya. (2023). https://www.analyticsvidhya.com/blog/2023/05/chatbot-evolution-chatgpt-vs-rule-based/#:~:text=The architecture of rule-based,user interacts with the chatbot.

[18] What are Chatbots, Codecademy. (2023). https://www.codecademy.com/article/what-are-chatbots.

[19] G. Bragg, How Do AI Chatbots Work: Exploring the Basics, WebiO. (2023). https://www.webio.com/blog/how-ai-chatbots-work.

[20] S. Hitter, What is a Large Language Model?, EWeek. (2023). https://www.eweek.com/artificial-intelligence/large-language-model/#:~:text=Large language models are created,large quantities of textual data.

[21] F.L. Somoye, How is ChatGPT trained?, PCguide. (2023). https://www.pcguide.com/apps/chat-gpt-trained/.

[22] A. Oppermann, What is Google's Bard?, Built In. (2023). https://builtin.com/artificial-intelligence/bard.

[23] M. Walsh, ChatGPT Statistics (2023)-- Key Facts and Figures, (2023).

[24] A. Gautam, Google Bard AI and ChatGPT: Which One is Right for You?, LinkedIn. (2023).

[25] What Is FPE? - SFPE, (n.d.). https://www.sfpe.org/career/what-is-fpe (accessed March 30, 2023).

[26] S.D. Christian, A Guide to Fire Safety Engineering, British Standards Institution, 2003.

[27] M. Dillon, L. Garland, Ancient Rome: A Sourcebook, Routledge, 2013.

[28] James Braidwood - Wikipedia, (n.d.). https://en.wikipedia.org/wiki/James_Braidwood

(accessed March 30, 2023).

[29] D.D. Drysdale, Fire Dynamics, Encycl. Phys. Sci. Technol. (2003) 869–892. https://doi.org/10.1016/B0-12-227410-5/00244-1.

[30] D. Drysdale, An Introduction to Fire Dynamics: Third Edition, 2011. https://doi.org/10.1002/9781119975465.

[31] V.K.R. Kodur, M.Z. Naser, Structural Fire Engineering, McGraw Hill, 2020.

[32] G.A. Khoury, Effect of fire on concrete and concrete structures, Prog. Struct. Eng. Mater. 2 (2000) 429–447. https://doi.org/https://doi.org/10.1002/pse.51.

[33] J.C.C. Liu, K.H. Tan, Y. Yao, A new perspective on the nature of fire-induced spalling in concrete, Constr. Build. Mater. 184 (2018) 581–590. https://doi.org/https://doi.org/10.1016/j.conbuildmat.2018.06.204.

[34] A.H. Buchanan, Structural Design for Fire Safety, 2nd Editio, John Wiley & Sons, 2017.

[35] V. Kodur, M.Z. Naser, Structural Fire Engineering, McGraw Hill Professional, 2020.

[36] V.K.R. Kodur, D.H. MacKinnon, Design of Concrete-Filled Hollow Structural Steel Columns for Fire Endurance, Eng. Journal, Am. Inst. Steel Constr. 37 (n.d.) 13–24.

[37] Eurocode CE. 2: Design of Concrete Structures--Part 1-2: General Rules--Structural Fire Design, London, U.K., 2004.

[38] C. McIntosh, EVACUATION | definition in the Cambridge English Dictionary, Cambridge Adv. Learn. Dict. Thes. (2023).

[39] M.J. Hurley, D. Gottuk, J.R. Hall, K. Harada, E. Kuligowski, M. Puchovsky, J. Torero, Jj.M. Watts, C. Wieczorek, SFPE handbook of fire protection engineering, fifth edition, 2016. https://doi.org/10.1007/978-1-4939-2565-0.

[40] G. Proulx, A stress model for people facing a fire, J. Environ. Psychol. (1993). https://doi.org/10.1016/S0272-4944(05)80146-X.

[41] R.F. Fahy, G. Proulx, “Panic” and human behaviour in fire, Proc. 4th Int. Symp. Hum. Behav. Fire. (2009).

[42] S.W. Gilbert, D.T. Butry, Identifying Vulnerable Populations to Death and Injuries from Residential Fires, 2018.

[43] A. Rahouti, R. Lovreglio, S. Gwynne, P. Jackson, S. Datoussaïd, A. Hunt, Human behaviour during a healthcare facility evacuation drills: Investigation of pre-evacuation and travel phases, Saf. Sci. 129 (2020) 104754. https://doi.org/10.1016/J.SSCI.2020.104754.

[44] H. Hostetter, M.Z. Naser, Characterizing disability in fire evacuation: A progressive review, J. Build. Eng. 53 (2022) 104573. https://doi.org/10.1016/J.JOBE.2022.104573.

[45] J.L. Bryan, A review of the examination and analysis of the dynamics of human behavior in the fire at the MGM Grand Hotel, Clark County, Nevada as determined from a selected questionnaire population, Fire Saf. J. 5 (1983) 233–240. https://doi.org/10.1016/0379-

7112(83)90021-8.

[46] J.D. Averill, D.S. Mileti, R.D. Peacock, E.D. Kuligowski, N. Groner, G. Proulx, P.A. Reneke, H.E. Nelson, Federal building and fire safety investigation of the World Trade Center Disaster: Occupant behavior, egress, and emergency communications, Nist Ncstar 1-7. (2005) 1–298.

[47] E.D. Kuligowski, Computer Evacuation Models for Buildings, in: SFPE Handb. Fire Prot. Eng., Springer New York, New York, NY, 2016: pp. 2152–2180. https://doi.org/10.1007/978-1-4939-2565-0_60.

[48] E.D. Kuligowski, R.D. Peacock, B.L. Hoskins, A review of building evacuation models, 2nd edition, technical note (NIST TN), Gaithersburg, MD, 2010.

[49] S. Gwynne, M. Amos, M. Kinateder, N. Bénichou, K. Boyce, C. Natalie van der Wal, E. Ronchi, The future of evacuation drills: Assessing and enhancing evacuee performance, Saf. Sci. (2020). https://doi.org/10.1016/j.ssci.2020.104767.

[50] Evacuation for persons with disabilities , (2023). https://emergency.vanderbilt.edu/vu/quick-ref-guides/evacuation-disabled.php/ (accessed March 30, 2023).

[51] K. Butler, E. Kuligowski, S. Furman, R. Peacock, Perspectives of occupants with mobility impairments on evacuation methods for use during fire emergencies, Fire Saf. J. 91 (2017) 955–963. https://doi.org/10.1016/J.FIRESAF.2017.04.025.

[52] D.P. Bliss, V. President, F. Operations, Creating the Research Roadmap for Smart Fire Fighting, NIST Spec. Publ. 1191. (2015) 1–247. https://doi.org/10.6028/NIST.SP.1191.

[53] Y. Zhang, X. Zhang, X. Huang, Design a safe firefighting time (SFT) for major fire disaster emergency response, Int. J. Disaster Risk Reduct. 88 (2023) 103606. https://doi.org/10.1016/J.IJDRR.2023.103606.

[54] M.Z. Naser, Mechanistically Informed Machine Learning and Artificial Intelligence in Fire Engineering and Sciences, Fire Technol. 57 (2021) 2741–2784. https://doi.org/10.1007/S10694-020-01069-8.

[55] ChatGPT Will Now Have Access To Real-Time Info From Bing Search, (n.d.). https://www.forbes.com/sites/katherinehamilton/2023/05/23/chatgpt-will-now-have-access-to-real-time-info-from-bing-search-report-says/?sh=7c2732d32397 (accessed June 16, 2023).

[56] Y. Huang, A. Gomaa, S. Semrau, M. Haderlein, S. Lettmaier, T. Weissmann, J. Grigo, H.B. Tkhayat, B. Frey, U.S. Gaipl, L.V. Distel, A. Maier, R. Fietkau, C. Bert, F. Putz, Benchmarking ChatGPT-4 on ACR Radiation Oncology In-Training (TXIT) Exam and Red Journal Gray Zone Cases: Potentials and Challenges for AI-Assisted Medical Education and Decision Making in Radiation Oncology, Cornell Univ. Med. Phys. (2023). https://doi.org/10.48550/arXiv.2304.11957.

[57] L. Passby, N. Jenko, A. Wernham, Performance of ChatGPT on dermatology Specialty Certificate Examination multiple choice questions, Clin. Exp. Dermatol. (2023). https://doi.org/10.1093/ced/llad197.